\documentclass[conference]{IEEEtran}
\usepackage{times}

\usepackage[numbers]{natbib}
\usepackage{multicol}
\usepackage[bookmarks=true]{hyperref}
\usepackage{graphicx}

\usepackage{multirow}
\usepackage{multicol}
\usepackage{graphicx}
\usepackage{xspace}
\usepackage{xcolor}
\usepackage{caption}
\let\labelindent\relax
\usepackage{enumitem} 
\usepackage{wrapfig}
\usepackage{bbding}  
\usepackage{pifont}  

\usepackage{booktabs}
\usepackage{float}
\usepackage{amsmath}
\usepackage{amssymb}
\usepackage{dsfont}
\usepackage{bm}
\usepackage{makecell}
\usepackage{subcaption}

\usepackage[capitalise, nameinlink]{cleveref}

\newcommand{\beamdojo}[0]{\mbox{\textsc{BeamDojo}}\xspace}

\newcommand{\ci}[1]{\tiny{\textcolor{gray}{~($\pm #1$)}}}

\usepackage{colortbl}
\definecolor{ourcolor}{HTML}{99e0eb}
\definecolor{ourblue}{HTML}{27a2c3}

\definecolor{tablecolor}{HTML}{ccf2f5} 

\definecolor{tablecolor2}{HTML}{ffcdb4}
\definecolor{citecolor}{HTML}{fe7b5b}
\definecolor{grey}{rgb}{0.9, 0.9, 0.9}
\usepackage{amssymb}

\usepackage{listings}
\definecolor{gred}{rgb}{0.859,0.267,0.216}
\definecolor{ggreen}{rgb}{0.059,0.616,0.345}

\definecolor{deepblue}{HTML}{27a2c3}

\definecolor{deepred}{HTML}{fe7b5b}

\usepackage[font=small,labelfont=bf]{caption}

\definecolor{citecolor}{HTML}{faa700} 
\definecolor{lblue}{HTML}{ffb114} 
\definecolor{ogreen}{HTML}{2E7D32}
\definecolor{bred}{HTML}{BF360C}
\definecolor{newbrown}{HTML}{795548}

\hypersetup{
    colorlinks=true,
    linkcolor=orange,
    urlcolor=orange,
    citecolor=orange,
}

\begin{document}

\title{BeamDojo: Learning Agile Humanoid \\Locomotion on Sparse Footholds}

\author{\authorblockN{Huayi Wang\textsuperscript{1,2} \
\quad Zirui Wang\textsuperscript{1,3} \
\quad Junli Ren\textsuperscript{1,4} \
\quad Qingwei Ben\textsuperscript{1,5} \
\quad Tao Huang\textsuperscript{1,2} \ \\
\quad Weinan Zhang\textsuperscript{1,2,\dag} \ 
\quad Jiangmiao Pang\textsuperscript{1,\dag} \ 
}
\authorblockA{
\textsuperscript{1}Shanghai AI Laboratory \quad 
\textsuperscript{2}Shanghai Jiao Tong University \quad
\textsuperscript{3}Zhejiang University \quad \\
\textsuperscript{4}The University of Hong Kong \quad
\textsuperscript{5}The Chinese University of Hong Kong\\
\textsuperscript{\dag}Corresponding Authors.\vspace{0.05cm}\\
Website: \href{https://why618188.github.io/beamdojo}{\texttt{https://why618188.github.io/beamdojo}}
}
}



%

\twocolumn[{%
\renewcommand\twocolumn[1][]{#1}%
\maketitle
\vspace{-0.5cm}
\begin{center}
    \centering
    \captionsetup{type=figure}
     \includegraphics[width=1.0\textwidth]{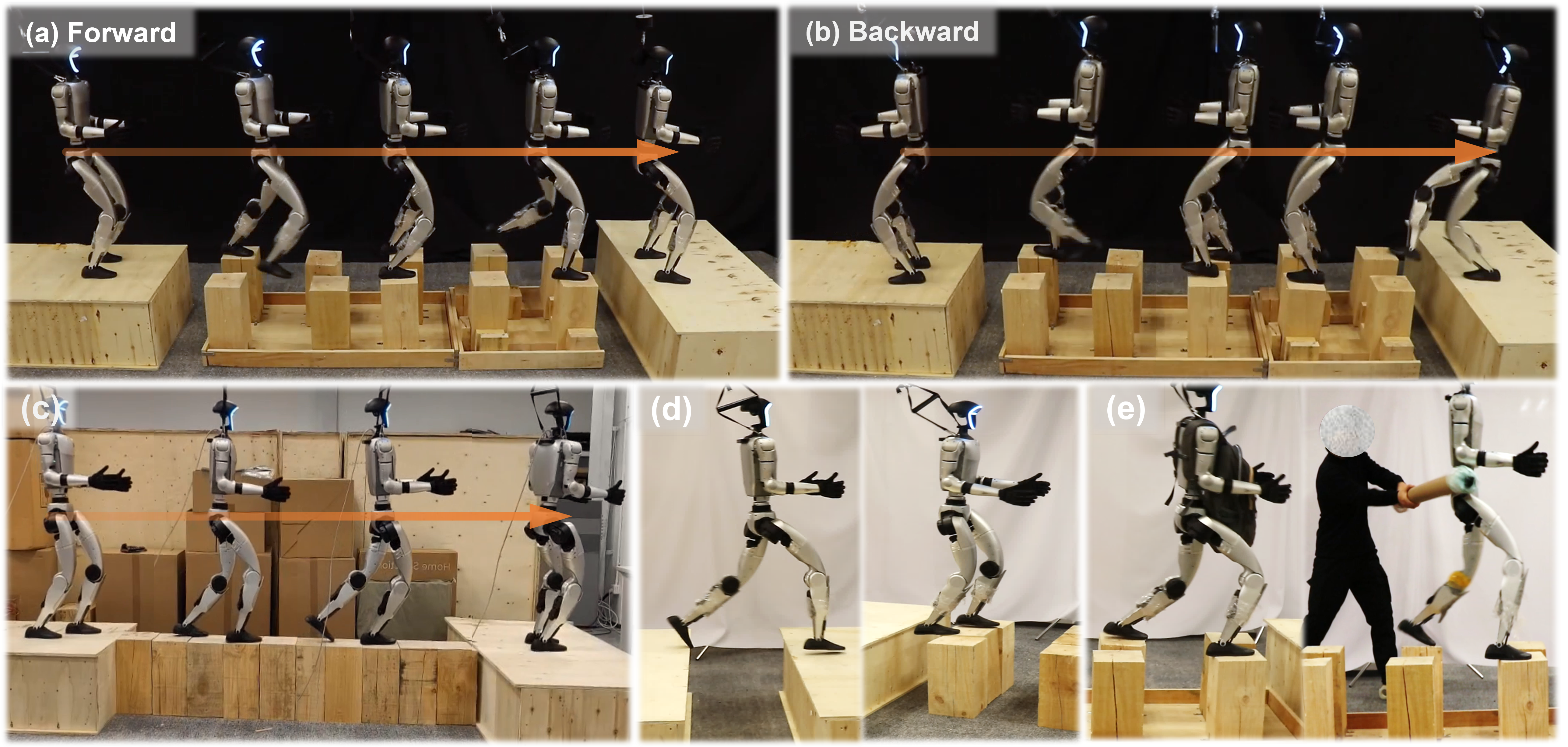}
     \vspace{-0.17in}
    \caption{
    Our proposed framework, \textbf{\beamdojo}, enables agile and robust humanoid locomotion across challenging sparse footholds. 
    \textbf{Top Row:} Utilizing global mapping information from LiDAR, the humanoid demonstrates the ability to precisely traverse both forward and backward on stepping stones, with arrows indicating the direction of movement.
    \textbf{Bottom Left:} The humanoid skillfully traverses a narrow balance beam.
    \textbf{Bottom Middle:} Despite being trained without exposure to gaps and balancing beams, the humanoid achieves zero-shot generalization to various sparse foothold terrains. 
    \textbf{Bottom Right:} Humanoid exhibits remarkable robustness, maintaining stable locomotion under external disturbances and additional payloads.
    }
    \label{fig:teaser}
\end{center}
\vspace{-0.0in}
}]

\begin{abstract}
Traversing risky terrains with sparse footholds poses a significant challenge for humanoid robots, requiring precise foot placements and stable locomotion. Existing learning-based approaches often struggle on such complex terrains due to sparse foothold rewards and inefficient learning processes. To address these challenges, we introduce \beamdojo, a reinforcement learning (RL) framework designed for enabling agile humanoid locomotion on sparse footholds. \beamdojo begins by introducing a sampling-based foothold reward tailored for polygonal feet, along with a double critic to balancing the learning process between dense locomotion rewards and sparse foothold rewards. To encourage sufficient trial-and-error exploration, \beamdojo incorporates a two-stage RL approach: the first stage relaxes the terrain dynamics by training the humanoid on flat terrain while providing it with task-terrain perceptive observations, and the second stage fine-tunes the policy on the actual task terrain. Moreover, we implement a onboard LiDAR-based elevation map to enable real-world deployment. Extensive simulation and real-world experiments demonstrate that \beamdojo achieves efficient learning in simulation and enables agile locomotion with precise foot placement on sparse footholds in the real world, maintaining a high success rate even under significant external disturbances.
\end{abstract}

\IEEEpeerreviewmaketitle

\section{Introduction}

Traversing risky terrains with sparse footholds, such as stepping stones and balancing beams, presents a significant challenge for legged locomotion. Achieving agile and safe locomotion on such environment requires robots to accurately process perceptive information, make precise footstep placement within safe areas, and maintain base stability throughout the process~\cite{yu2024walking, Zhang2023LearningAL}.

Existing works have effectively addressed this complex task for quadrupedal robots~\cite{gangapurwala2022rloc, grandia2023perceptive, jenelten2024dtc, lu2024learning, xie2022glide, yu2024walking, yu2021visual, Zhang2023LearningAL, zhu2024robust}. However, these methods encounter great challenges when applied to humanoid robots, primarily due to a key difference in foot geometry. Although the foot of most quadrupedal robots and some simplified bipedal robots~\cite{englsberger2011bipedal, kajita20013d} can be modeled as a point, the foot of humanoid robots is often represented as a polygon~\cite{chestnutt2005footstep, griffin2019footstep, hornung2012adaptive, stumpf2014supervised}. For traditional model-based methods, this requires additional half-space constraints defined by linear inequalities, which impose a significant computational burden for online planning~\cite{dai2016planning, deits2014footstep, griffin2019footstep, margolis2021learning, stumpf2014supervised}. In the case of reinforcement learning (RL) methods, foothold rewards designed for point-shaped feet are not suitable for evaluating foot placement of polygon-shaped feet~\cite{yu2024walking}. Hybrid methods, which combine RL with model-based controllers, face similar challenges in online planning for humanoid feet~\cite{gangapurwala2022rloc, jenelten2024dtc, xie2022glide}. Furthermore, the higher degrees of freedom and the inherently unstable morphology of humanoid robots make it even more difficult to achieve agile and stable locomotion over risky terrains.

On the other hand, recent advancements in learning-based humanoid robot locomotion have demonstrated impressive robustness across various tasks, including walking~\cite{chen2024learning, radosavovic2024learning, radosavovic2024humanoid, van2024revisiting}, stair climbing~\cite{cuiadapting, gu2024advancing, long2024learninghumanoid}, parkour~\cite{zhuang2024humanoid}, and whole-body control~\cite{cheng2024expressive, fu2024humanplus, he2024omnih2o, he2024learning, he2024hover, ji2024exbody2, jiang2024harmon}, etc. However, these methods still struggle with complex terrains and agile locomotion on fine-grained footholds. Enabling agile movement on risky terrains for humanoid robots presents several challenges. First, the reward signal for evaluating foot placement is sparse, typically provided only after completing a full sub-process (e.g., lifting and landing a foot), which makes it difficult to assign credit to specific states and actions~\cite{sutton1984temporal}. Second, the learning process is highly inefficient, as a single misstep often leads to early termination during training, hindering sufficient exploration. Additionally, obtaining reliable perceptual information is challenging due to sensory limitations and environmental noise~\cite{zhuang2024humanoid}.

In this work, we introduce \beamdojo, a novel reinforcement learning-based framework for controlling humanoid robots traversing risky terrains with sparse footholds. The name \beamdojo combines the words ``beam'' (referring to sparse footholds such as beams) and ``dojo'' (a place of training or learning), reflecting the goal of training agile locomotion on such challenging terrains. We begin by defining a sampling-based foothold reward, designed to evaluate the foot placement of a polygonal foot model. To address the challenge of sparse foothold reward learning, we propose using double critic architecture to separately learn the dense locomotion rewards and the sparse foothold reward. Unlike typical end-to-end RL methods~\cite{yu2024walking, Zhang2023LearningAL}, \beamdojo further incorporates a two-stage approach to encourage fully trial-and-error exploration. In the first stage, terrain dynamics constraints are relaxed, allowing the humanoid robot to practice walking on flat terrain while receiving perceptive information of the target task terrain (e.g., sparse beams), where missteps will incur a penalty but do not terminate the episode. In the second stage, the policy is fine-tuned on the true task terrain. To enable deployment in real-world scenarios, we further implement a LiDAR-based, robot-centric elevation map with carefully designed domain randomization in simulation training.

As shown in Fig.~\ref{fig:teaser}, \beamdojo skillfully enables humanoid robots to traverse risky terrains with sparse footholds, such as stepping stones and balancing beams. Through extensive simulations and real-world experiments, we demonstrate the efficient learning process of \beamdojo and its ability to achieve agile locomotion with precise foot placements in real-world scenarios.

The contributions of our work are summarized as follows: 
\begin{itemize}
    \item We propose \beamdojo, a two-stage RL framework that combines a newly designed foothold reward for the polygonal foot model and a double critic, enabling humanoid locomotion on sparse footholds. 
    \item We implement a LiDAR-based elevation map for real-world deployment, incorporating carefully designed domain randomization in simulation training.
    \item We conduct extensive experiments both in simulation and on Unitree G1 Humanoid, demonstrating agile and robust locomotion on sparse footholds, with a high zero-shot sim-to-real transfer success rate of 80\%.  To the best of our knowledge, \beamdojo is the first learning-based method to achieve fine-grained foothold control on risky terrains with sparse footholds.
\end{itemize}

\section{Related Works}

\subsection{Locomotion on Sparse Footholds}

Walking on sparse footholds has been a long-standing application of perceptive legged locomotion. Existing works often employs model-based hierarchical controllers, which decompose this complex task into separate stages of perception, planning, and control~\cite{grandia2023perceptive, griffin2019footstep, jenelten2020perceptive, mastalli2020motion, melon2021receding, mishra2024efficient, winkler2018gait}. However, model-based controllers react sensitively to violation of model assumptions, which hinders applications in real-world scenarios. Recent studies have explored combining RL with model-based controllers, such as using RL to generate trajectories that are then tracked by model-based controllers~\cite{gangapurwala2022rloc, yu2021visual, xie2022glide}, or employing RL policies to track trajectories generated by model-based planners~\cite{jenelten2024dtc}. While demonstrating remarkable performance, these decoupled architectures can constrain the adaptability and coordination of each module.

Subsequent works have explored end-to-end learning frameworks that train robots to walk on sparse footholds using perceptive locomotion controllers~\cite{agarwal2023legged, cheng2024extreme, yang2023neural, yu2024walking, Zhang2023LearningAL}. Despite their focus being limited to quadrupeds, a majority of these works rely on depth cameras for exteroceptive observations, which are limited by the camera's narrow field of view and restrict the robot to moving backward~\cite{agarwal2023legged, cheng2024extreme, yang2023neural, yu2024walking}. Additionally, an image processing module is often necessary to bridge the sim-to-real gap between the captured depth images and the terrain heightmap used during training~\cite{agarwal2023legged, cheng2024extreme, yang2023neural, yu2024walking}.

In contrast to the aforementioned literature, this work achieves agile humanoid locomotion over risky terrains that addressing unique challenges specific to humanoid systems, such as foot geometry. Additionally, we implement a LiDAR-based elevation map to enhance the task, demonstrating that the robot can move smoothly both forward and backward using the robotics-centric elevation map as the perception module.

\subsection{Reinforcement Learning in Locomotion Control}

Reinforcement learning has been widely applied in legged locomotion control~\cite{cheng2024extreme, he2024agile, kumar2021rma, lee2020learning, long2024hybrid, margolis2024rapid, miki2022learning, nahrendra2023dreamwaq, zhuang2023robot}, benefiting from the policy update stability and high data efficiency provided by Proximal Policy Optimization (PPO)~\cite{schulman2017proximal}. To adapt learned policies to diverse target tasks and ensure hardware deployability, previous works have designed two-stage training frameworks that aim to bridge the sim-to-real gap in the observation space~\cite{kumar2021rma, lee2020learning}. In contrast, this work introduces a novel two-stage training approach specifically aimed at improving sample efficiency, particularly addressing the challenge of early termination when walking on sparse terrains. This design not only enhances performance but also ensures more efficient learning in complex, real-world environments.

\section{Problem Formulation}

This work aims to develop an terrain-aware humanoid locomotion policy, where controllers are trained via reinforcement learning (RL). The RL problem is formulated as a Markov Decision Process (MDP) $\mathcal{M} = \langle \mathcal{S}, \mathcal{A}, T, \mathcal{O}, r, \gamma \rangle$, where $\mathcal{S}$ and $\mathcal{A}$ denote the state and action spaces, respectively. The transition dynamics are represented by $T(s' \mid\! s, a)$, the reward function by $r(s, a)$, and the discount factor by $\gamma \in [0, 1]$. The primary objective is to optimize the policy $\pi(a_t \mid s_t)$ to maximize the discounted cumulative rewards: \begin{equation} \max_\pi J(\mathcal{M}, \pi) = \mathbb{E}\left[ \sum_{t=0}^\infty \gamma^t r(s_t, a_t)\right]. \end{equation}

In this work, the agent only has access to partial observations $\mathbf{o} \in \mathcal{O}$ due to sensory limitations and environmental noise, which provide incomplete information about the true state. Consequently, the agent functions within the framework of a Partially Observable Markov Decision Process (POMDP).

\section{Methods}

\subsection{Foothold Reward}
\label{sec:method_footholdreward}

\begin{figure}[t]
    \centering
    \includegraphics[width=0.85\linewidth]{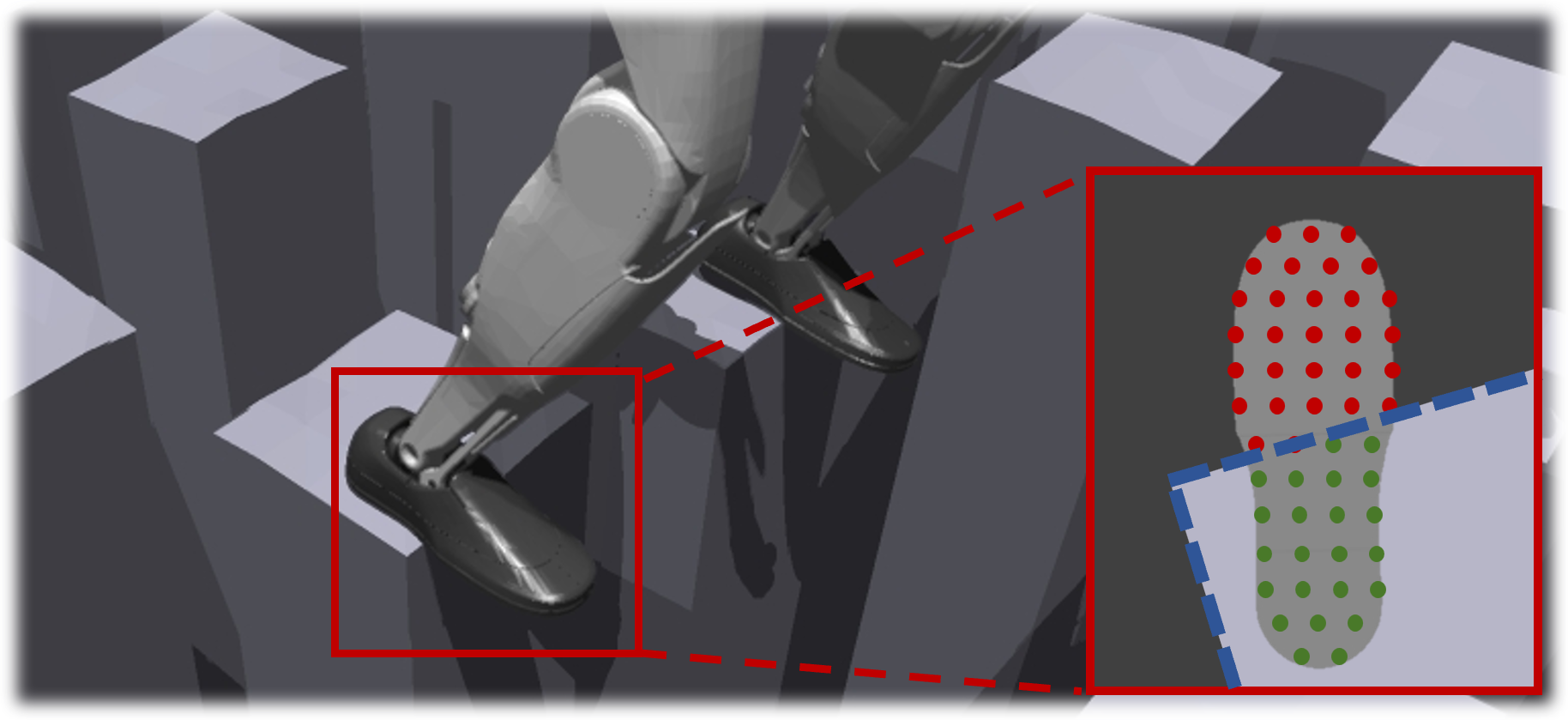}
    \caption{\textbf{Foothold Reward.} We sample $n$ points under the foot. Green points indicate contact with the surface within the safe region, while red points represent those not in contact with the surface.}
    \label{fig:foothold_reward}
\end{figure}

To accommodate the polygonal foot model of the humanoid robot, we introduce a sampling-based foothold reward that evaluates foot placement on sparse footholds.This evaluation is determined by the overlap between the foot's placement and designated safe areas, such as stones and beams. Specifically, we sample $n$ points on the soles of the robot’s feet, as illustrated in Fig.~\ref{fig:foothold_reward}. For each $j$-th sample on foot $i$, let $d_{ij}$ denotes the global terrain height at the corresponding position. The penalty foothold reward $r_\text{foothold}$ is defined as:
\begin{equation}
    r_\text{foothold} = -\sum_{i=1}^2 \mathbb{C}_i \sum_{j=1}^n \mathds{1} \{ d_{ij} < \epsilon \},
\end{equation}
where $\mathbb{C}_i$ is an indicator function that specifies whether foot $i$ is in contact with the terrain surface, and $\mathds{1}$ is the indicator function for a condition. The threshold $\epsilon$ is a predefined depth tolerance, and when $d_{ij} < \epsilon$, it indicates that the terrain height at this sample point is significantly low, implying improper foot placement outside of a safe area. This reward function encourages the humanoid robot to maximize the overlap between its foot placement and the safe footholds, thereby improving its terrain-awareness capabilities.

\begin{figure*}[t]
    \centering
    \includegraphics[width=\linewidth]{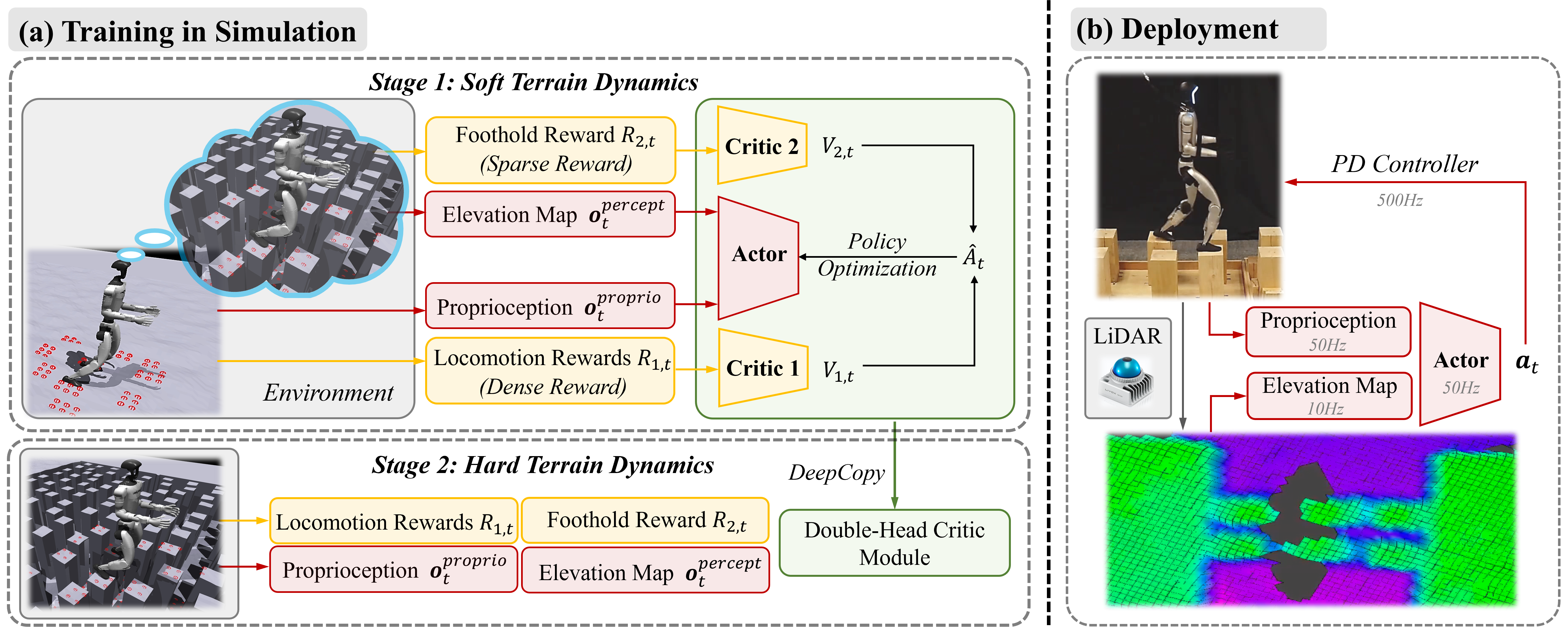}
    \caption{\textbf{Overview of \beamdojo.} (a) \textbf{Training in Simulation:} In stage 1, proprioceptive and perceptive information, locomotion rewards and the foothold reward are decoupled respectively, with the former obtained from flat terrain and the latter from task terrain. The double critic module separately learns two reward groups. In stage 2, the policy is fine-tuned on the task terrain, utilizing the full set of observations and rewards. (b) \textbf{Real-world deployment:} The robot-centric elevation map, reconstructed using LiDAR data, is combined with proprioceptive information to serve as the input for the actor.}
    \label{fig:framework}
\end{figure*}

\subsection{Double Critic for Sparse Reward Learning}
\label{sec:method_doublecritic}

The task-specific foothold reward $r_\text{foothold}$ is a sparse reward. To effectively optimize the policy, it is crucial to carefully balance this sparse reward with dense locomotion rewards which are crucial for gait regularization~\cite{zargarbashi2024robotkeyframing}. Inspired by~\cite{huang2022reward, xu2023composite, zargarbashi2024robotkeyframing}, we adopt a double critic framework based on PPO, which effectively decouples the mixture of dense and sparse rewards. 

In this framework, we train two separate critic networks, $\left\{V_{\phi_1}, V_{\phi_2}\right\}$, to independently estimate value functions for two distinct reward groups: (i) the regular locomotion reward group (dense rewards), $R_1=\{r_i\}_{i=0}^n$, which have been studied in quadruped locomotion tasks~\cite{margolis2023walk} and humanoid locomotion tasks~\cite{long2024learninghumanoid}, and (ii) the task-specific foothold reward group (sparse reward), $R_2=\{r_\text{foothold}\}$.

The double critic process is illustrated in Fig.~\ref{fig:framework}. Specifically, each value network $V_{\phi_i}$ is updated independently for its corresponding reward group $R_i$ with temporal difference loss (TD-loss):
\begin{equation}
    \mathcal{L}(\phi_i)=\mathbb{E} \left[ \left\| R_{i, t} + \gamma V_{\phi_i}(s_{t+1}) - V_{\phi_i}(s_t) \right\|^2 \right],
\end{equation}
where $\gamma$ is the discount factor. Then the respective advantages $\{\hat{A}_{i, t}\}$ are calculated using Generalized Advantage Estimation (GAE)~\cite{schulman2015high}:
\begin{gather}
    \delta_{i, t} = R_{i, t} + \gamma V_{\phi_i}(s_{t+1}) - V_{\phi_i}(s_t), \\
    \hat{A}_{i, t} = \sum_{l=0}^{\infty} (\gamma \lambda)^l \delta_{i, t+l},
\end{gather}
where $\lambda$ is the balancing parameter. These advantages are then individually normalized and synthesized into an overall advantage:
\begin{equation}
   \hat{A}_t = w_1 \cdot \frac{\hat{A}_{1, t} - \mu_{\hat{A}_{1, t}}}{\sigma_{\hat{A}_{1, t}}} + w_2 \cdot \frac{\hat{A}_{2, t} - \mu_{\hat{A}_{2, t}}}{\sigma_{\hat{A}_{2, t}}} ,
\end{equation}
where $w_i$ is the weight for each advantage component, and $\mu_{\hat{A}_{i, t}}$ and $\sigma_{\hat{A}_{i, t}}$ are the batch mean and standard deviation of each component. This overall advantage is then used to update the policy:
\begin{equation}
    \mathcal{L}(\theta) = \mathbb{E} \left[ \min \left( \alpha_t (\theta)\hat{A}_t, \text{clip}(\alpha_t(\theta), 1-\epsilon, 1+\epsilon)\hat{A}_t \right) \right],
\end{equation}
where $\alpha_t(\theta)$ is the probability ratio, and $\epsilon$ is the clipping hyperparameter.

This double critic design provides a modular, plug-and-play solution for handling specialized tasks with sparse rewards, while effectively addressing the disparity in reward feedback frequencies within a mixed dense-sparse environment~\cite{zargarbashi2024robotkeyframing}. The detailed reward terms are provided in Appendix~\ref{sec:append_reward}.

\subsection{Learning Terrain-Aware Locomotion via Two-Stage RL}
\label{sec:method_twostage}

To address the early termination problem in complex terrain dynamics and encourage full trial-and-error exploration, we design a novel two-stage reinforcement learning (RL) approach for terrain-aware locomotion in simulation. As illustrated in Fig.~\ref{fig:framework}, in the first stage, termed the ``soft terrain dynamics constraints'' phase, the humanoid robot is trained on flat terrain while being provided with a corresponding height map of the true task terrains (e.g., stepping stones). This setup encourages broad exploration without the risk of early termination from missteps. Missteps are penalized but do not lead to termination, allowing the humanoid robot to develop foundational skills for terrain-aware locomotion. In the second stage, termed the ``hard terrain dynamics constraints'' phase, we continue training the humanoid on the real terrains in simulation, where missteps result in termination. This stage fine-tunes the robot's ability to step on challenging terrains accurately.

\subsubsection{Stage 1: Soft Terrain Dynamics Constraints Learning}

In this stage, we first map each task terrain (denoted as $\mathcal{T}$) to a flat terrain (denoted as $\mathcal{F}$) of the same size. Both terrains share the same terrain noise, with points are one-to-one corresponding. The only difference is that the flat terrain $\mathcal{F}$ fills the gaps in the real terrain $\mathcal{T}$. 

We let the humanoid robot traverse the terrain $\mathcal{F}$, receiving proprioceptive observations, while providing perceptual feedback in the form of the elevation map of terrain $\mathcal{T}$ at the corresponding humanoid's base position. This setup allows the robot to ``imagine'' walking on the true task terrain while actually traversing the safer flat terrain, where missteps do not lead to termination. To expose the robot to real terrain dynamics, we use the foothold reward (introduced in Section~\ref{sec:method_footholdreward}). In this phase, this reward is provided by the terrain $\mathcal{T}$, where $d_{ij}$ is the height of the true terrain at the sampling point, while locomotion rewards are provided by the terrain $\mathcal{F}$. 

This design successfully decouples the standard locomotion task and the task of traversing sparse footholds: flat terrain, $\mathcal{F}$, provides proprioceptive information and locomotion rewards to learn regular gaits, while risky task terrain, $\mathcal{T}$, offers perceptive information and the foothold reward to develop terrain-awareness skills. We train these two reward components separately using a double critic framework, as described in Section~\ref{sec:method_doublecritic}. 

Furthermore, by allowing the humanoid robot to traverse the flat terrain while applying penalties for missteps instead of terminating the episode, the robot can continuously attempt foothold placements, making it much easier to obtain successful positive samples. In contrast, conventional early termination disrupts entire trajectories, making it extremely difficult to acquire safe foothold samples when learning from scratch. This approach significantly improves sampling efficiency and alleviates the challenges of exploring terrains with sparse footholds.

\subsubsection{Stage 2: Hard Terrain Dynamics Constraints Learning}

In the second stage, we fine-tune the policy directly on the task terrain $\mathcal{T}$. Unlike in Stage 1, missteps on $\mathcal{T}$ now result in immediate termination. This enforces strict adherence to the true terrain constraints, requiring the robot to develop precise and safe locomotion strategies.

To maintain a smooth gait and accurate foot placements, we continue leveraging the double-critic framework to optimize both locomotion rewards and the foothold reward $r_\text{foothold}$ Here, $d_{ij}$ again represents the height of terrain $\mathcal{T}$ at the given sampling point.

\subsection{Training in Simulation}
\subsubsection{Observation Space and Action Space}

The policy observations, denoted as $\mathbf{o}_t$, consist of four components:
\begin{equation} 
    \mathbf{o}_t = \left[\mathbf{c}_t, \mathbf{o}^\text{proprio}_t, \mathbf{o}_t^\text{percept}, \mathbf{a}_{t-1} \right]. 
\end{equation}

The commands $\mathbf{c}_t \in \mathbb{R}^3$ specify the desired velocity, represented as $\left[ \mathbf{v}_x^c, \mathbf{v}_y^c, \boldsymbol{\omega}_\text{yaw}^c \right]$. These denote the linear velocities in the longitudinal and lateral directions, and the angular velocity in the horizontal plane, respectively. The proprioceptive observations $\mathbf{o}_t^\text{proprio} \in \mathbb{R}^{64}$ include the base angular velocity $\bm{\omega}_t \in \mathbb{R}^3$, gravity direction in the robot's frame $\mathbf{g}_t \in \mathbb{R}^3$, joint positions $\bm{\theta}_t \in \mathbb{R}^{29}$, and joint velocities $\dot{\bm{\theta}}_t \in \mathbb{R}^{29}$. The perceptive observations $\mathbf{o}_t^\text{percept} \in \mathbb{R}^{15 \times 15}$ correspond to an egocentric elevation map centered around the robot. This map samples $15 \times 15$ points within a $0.1$ m grid in both the longitudinal and lateral directions. The action of last timestep $\mathbf{a}_{t-1} \in \mathbb{R}^{12}$ is also included to provide temporal context.

The action $\mathbf{a}_t \in \mathbb{R}^{12}$ represents the target joint positions for the $12$ lower-body joints of the humanoid robot, which are directly output by the actor network. For the upper body joints, the default position is used for simplicity. A proportional-derivative (PD) controller converts these joint targets into torques to track the desired positions.

\subsubsection{Terrain and Curriculum Design}
\label{sec:method_curriculum}

\begin{figure}[t]
    \centering
    \includegraphics[width=\linewidth]{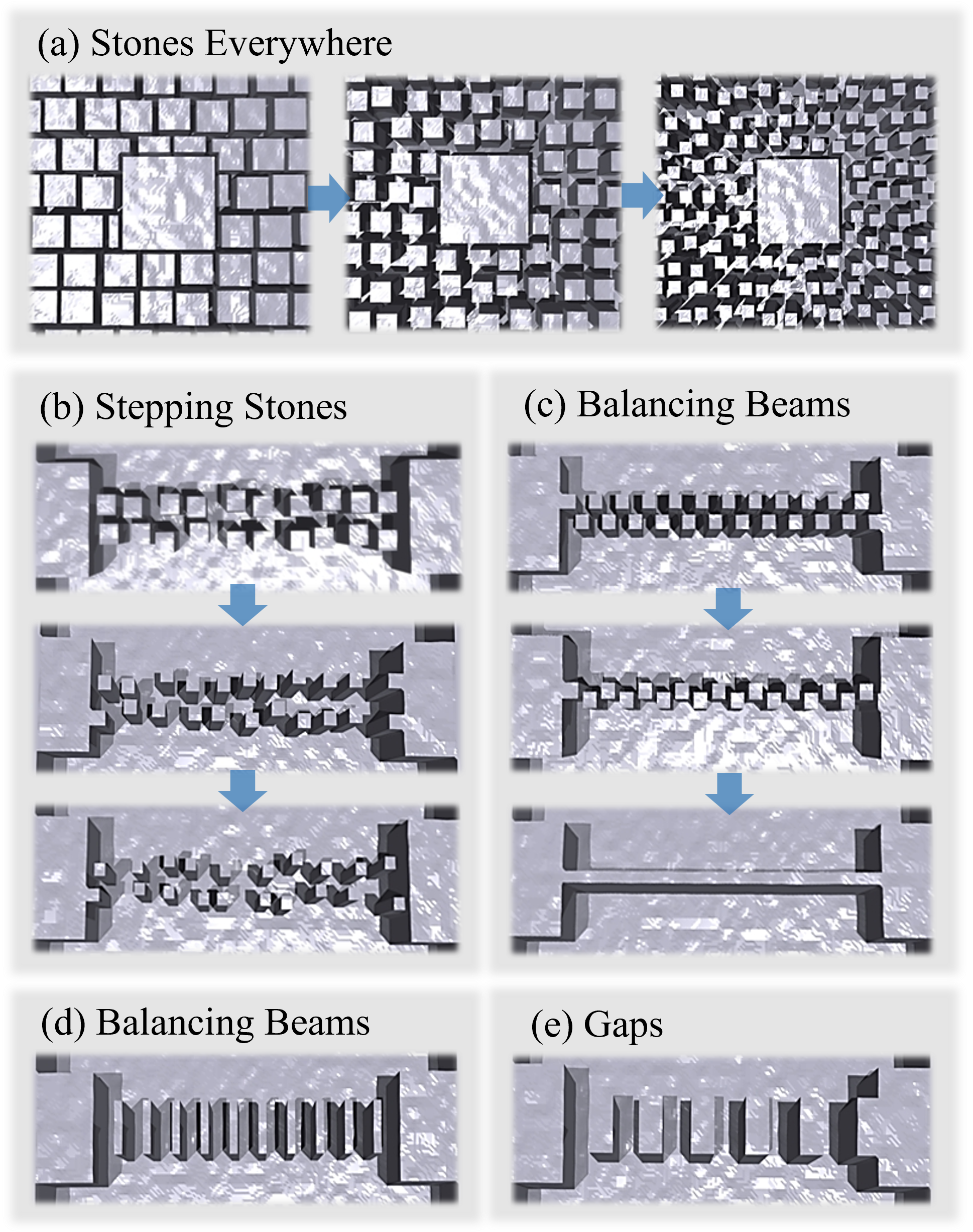}
    \caption{\textbf{Terrain Setting in Simulation.} (a) is used for stage 1 training, while (b) and (c) are used for stage 2 training. The training terrain progression is listed from simple to difficult. (b)-(e) are used for evaluation.}
    \label{fig:terrain}
\end{figure}

Inspired by~\cite{heess2017emergence, yu2024walking, Zhang2023LearningAL}, we design five types of sparse foothold terrains for the two-stage training and evaluation:

\begin{itemize}
    \item \textit{Stones Everywhere}: This is a general sparse foothold terrain where stones are scattered across the entire terrain. The center of the terrain is a platform surrounded by stones, as shown in Fig.~\ref{fig:terrain}(a). The stones are uniformly distributed within sub-square grids. As the curriculum progresses, the stone size decreases and the sparsity increases.
    
    \item \textit{Stepping Stones}: This terrain consists of two lines of stepping stones in the longitudinal direction, connected by two platforms at each end, as shown in Fig.~\ref{fig:terrain}(b). Each stone is uniformly distributed in two sub-square grids, with the same curriculum effect as in \textit{Stones Everywhere}.
    
    \item \textit{Balancing Beams}: In the initial curriculum level, this terrain has two lines of separate stones in the longitudinal direction. As the curriculum progresses, the size of the stones decreases and their lateral distance reduces, eventually forming a single line of balancing beams, as shown in Fig.~\ref{fig:terrain}(c). This terrain is challenging for the robot as it must learn to keep its feet together on the beams without colliding with each other, while maintaining the center of mass. This requires a distinct gait compared to regular locomotion tasks.
    
    \item \textit{Stepping Beams}: This terrain consists of a sequence of beams to step on, randomly distributed along the longitudinal direction, with two platforms at either end, as illustrated in Fig.~\ref{fig:terrain}(d). This terrain, along with the \textit{Stones Everywhere} and \textit{Stepping Stones} terrains, requires the robot to place its footholds with high precision.
    
    \item \textit{Gaps}: This terrain consists of several gaps with random distances between them, as shown in Fig.~\ref{fig:terrain}(e). This terrain requires the robot to make large steps to cross the gaps.
\end{itemize}

We begin by training the robot on the \textit{Stones Everywhere} terrain in Stage 1 with soft terrain constraints to develop a generalizable policy. In Stage 2, the policy is fine-tuned on the \textit{Stepping Stones} and \textit{Balancing Beams} terrains with hard terrain constraints. The commands used in these two stages are detailed in Table~\ref{tab:command}. Note that in Stage 2, only a single x-direction command is given, with no yaw command provided. This means that if the robot deviates from facing forward, no correction command is applied. We aim for the robot to learn to consistently face forward from preceptive observation, rather than relying on continuous yaw corrections.

\begin{table}[t]
    \centering
    \caption{Commands Sampled in Two Stage RL Training}
    \begin{tabular}{lll}
    \toprule[1.0pt]
    \textbf{Term} & \textbf{Value (stage 1)} & \textbf{Value (stage 2)} \\
    
    \midrule[1.pt]
    $\mathbf{v}_x^c$ & $\mathcal{U}(-1.0, 1.0)$ m/s & $\mathcal{U}(-1.0, 1.0)$ m/s \\ [0.2ex]
    $\mathbf{v}_y^c$ & $\mathcal{U}(-1.0, 1.0)$ m/s & $\mathcal{U}(0.0, 0.0)$ m/s \\ [0.2ex]
    $\boldsymbol{\omega}_\text{yaw}^c$ & $\mathcal{U}(-1.0, 1.0)$ rad/s & $\mathcal{U}(0.0, 0.0)$ m/s \\ [0.2ex]    
    \bottomrule[1.0pt]
    \end{tabular}
    \label{tab:command}
\end{table}

For evaluation, the \textit{Stepping Stones}, \textit{Balancing Beams}, \textit{Stepping Beams}, and \textit{Gaps} terrains are employed. Remarkably, our method demonstrates strong zero-shot transfer capabilities on the latter two terrains, despite the robot being trained exclusively on the first three terrains.

The curriculum is designed as follows: the robot progresses to the next terrain level when it successfully traverses the current terrain level three times in a row. Furthermore, the robot will not be sent back to an easier terrain level before it pass all levels, as training on higher-difficulty terrains is challenging at first. The detailed settings of the terrain curriculum are presented in the Appendix~\ref{sec:append_curriculum}.

\subsubsection{Sim-to-Real Transfer}
\label{sec:sim2real}

To enhance robustness and facilitate sim-to-real transfer, we employ extensive domain randomization~\cite{tobin2017domain, xie2021dynamics} on key dynamic parameters. Noise is injected into observations, humanoid physical properties, and terrain dynamics. Additionally, to address the large sim-to-real gap between the ground-truth elevation map in simulation and the LiDAR-generated map in reality—caused by factors such as odometry inaccuracies, noise, and jitter—we introduce four types of elevation map measurement noise during height sampling in the simulator:
\begin{itemize}
    \item \textit{Vertical Measurement}: Random vertical offsets are applied to the heights for an episode, along with uniformly sampled vertical noise added to each height sample at every time step, simulating the noisy vertical measurement of the LiDAR.
    \item \textit{Map Rotation}: To simulate odometry inaccuracies, we rotate the map in roll, pitch, and yaw. For yaw rotation, we first sample a random yaw noise. The elevation map, initially aligned with the robot's current orientation, is then resampled by adding the yaw noise, resulting in a new elevation map corresponding to the updated orientation. For roll and pitch rotations, we randomly sample the biases $\left[ h_x, h_y \right]$ and perform linear interpolation from $-h_x$ to $h_x$ along the $x$-direction and from $-h_y$ to $h_y$ along the $y$-direction. The resulting vertical height map noise is then added to the original elevation map.
    \item \textit{Foothold Extension}: Random foothold points adjacent to valid footholds are extended, turning them into valid footholds. This simulates the smoothing effect that occurs during processing of LiDAR elevation data.
    \item \textit{Map Repeat}: To simulate delays in elevation map updates, we randomly repeat the map from the previous timestep.
\end{itemize}

The detailed domain randomization settings are provided in Appendix~\ref{sec:append_random}.

\subsection{Real-world Deployment}

\subsubsection{Hardware Setup}

We use Unitree G1 humanoid robot for our experiments in this work. The robot weighs 35 kg, stands 1.32 meters tall, and features 23 actuated degrees of freedom: 6 in each leg, 5 in each arm, and 1 in the waist. It is equipped with a Jetson Orin NX for onboard computation and a Livox Mid-360 LiDAR, which provides both IMU data and feature points for perception.

\subsubsection{Elevation Map and System Design}

The raw point cloud data obtained directly from the LiDAR suffers from significant occlusion and noise, making it challenging to use directly. To address this, we followed~\cite{long2024learninghumanoid} to construct a robot-centric, complete, and robust elevation map. Specifically, we employed Fast LiDAR-Inertial Odometry (FAST-LIO)~\cite{xu2021fast, xu2022fast} to fuse LiDAR feature points with IMU data provided by the LiDAR. This fusion generates precise odometry outputs, which are further processed using robot-centric elevation mapping methods~\cite{Fankhauser2014RobotCentricElevationMapping, Fankhauser2018ProbabilisticTerrainMapping} to produce a grid-based representation of ground heights.

During deployment, the elevation map publishes information at a frequency of $10$ Hz, while the learned policy operates at $50$ Hz. The policy’s action outputs are subsequently sent to a PD controller, which runs at $500$ Hz, ensuring smooth and precise actuation.

\section{Experiments}
\subsection{Experimental Setup}

\setlength{\tabcolsep}{4pt}
\begin{table*}[!ht]
\caption{Benchmarked Comparison in Simulation.}
\label{tab:all-results}
\begin{center}
\begin{tabular}{lcccccccc}
\toprule[1.0pt]
\multicolumn{1}{l}{\multirow{2}{*}{}} & \multicolumn{2}{c}{Stepping Stones} & \multicolumn{2}{c}{Balancing Beams} & \multicolumn{2}{c}{Stepping Beams} &
\multicolumn{2}{c}{Gaps} \\
\cmidrule[\heavyrulewidth](lr){2-3} \cmidrule[\heavyrulewidth](lr){4-5} \cmidrule[\heavyrulewidth](lr){6-7} \cmidrule[\heavyrulewidth](lr){8-9}

\multicolumn{1}{l}{} & $R_{\mathrm{succ}}$ ($\%, \uparrow$) & $R_\mathrm{trav}$ ($\%, \uparrow$) & $R_\mathrm{succ}$ ($\%, \uparrow$) & $R_\mathrm{trav}$ ($\%, \uparrow$) & $R_\mathrm{succ}$ ($\%, \uparrow$) & $R_\mathrm{trav}$ ($\%, \uparrow$) & $R_\mathrm{succ}$ ($\%, \uparrow$) & $R_\mathrm{trav}$ ($\%, \uparrow$) \\
\midrule[0.8pt]
\rowcolor[gray]{0.9} \multicolumn{9}{l}{\textbf{\textit{Medium Terrain Difficulty}}} \\
\midrule[0.8pt]

PIM  & $71.00$\ci{1.53} & $78.29$\ci{2.49} & $74.67$\ci{2.08} & $82.19$\ci{4.96} & $88.33$\ci{3.61} & $93.16$\ci{4.78} & $\mathbf{98.00}$\ci{0.57} & $99.16$\ci{0.75} \\  [0.4ex]

Naive  & $48.33$\ci{6.11} & $47.79$\ci{5.76} & $57.00$\ci{7.81} & $71.59$\ci{8.14} & $92.00$\ci{2.52} & $92.67$\ci{3.62} & $95.33$\ci{1.53} & $98.41$\ci{0.67} \\  [0.4ex]

Ours w/o Soft Dyn  & $65.33$\ci{2.08} & $74.62$\ci{1.37} & $79.00$\ci{2.64} & $82.67$\ci{2.92} & $\mathbf{98.67}$\ci{2.31} & $\mathbf{99.64}$\ci{0.62} & $96.33$\ci{1.53} & $98.60$\ci{1.15} \\  [0.4ex]

Ours w/o Double Critic  & $83.00$\ci{2.00} & $86.64$\ci{1.96} & $88.67$\ci{2.65} & $90.21$\ci{1.95} & $96.33$\ci{1.15} & $98.88$\ci{1.21} & $\mathbf{98.00}$\ci{1.00} & $\mathbf{99.33}$\ci{0.38} \\  [0.4ex]

\textbf{\beamdojo}  & $\mathbf{95.67}$\ci{1.53} & $\mathbf{96.11}$\ci{1.22} & $\mathbf{98.00}$\ci{2.00} & $\mathbf{99.91}$\ci{0.07} & $98.33$\ci{1.15} & $99.28$\ci{0.65} & $\mathbf{98.00}$\ci{2.65} & $99.21$\ci{1.24} \\  

\midrule[0.8pt]

\rowcolor[gray]{0.9} \multicolumn{9}{l}{\textit{\textbf{Hard Terrain Difficulty}}} \\

\midrule[0.8pt]

PIM  & $46.67$\ci{2.31} & $52.88$\ci{2.86} & $33.00$\ci{2.31} & $45.2$8\ci{3.64} & $82.67$\ci{2.31} & $90.68$\ci{1.79} & $\mathbf{96.00}$\ci{1.00} & $\mathbf{98.27}$\ci{3.96} \\  [0.4ex]

Naive  & $00.33$\ci{0.57} & $21.17$\ci{1.71} & $00.67$\ci{1.15} & $36.25$\ci{7.85} & $82.00$\ci{3.61} & $88.91$\ci{3.75} & $31.00$\ci{3.61} & $62.70$\ci{4.08} \\  [0.4ex]

Ours w/o Soft Dyn  & $42.00$\ci{6.56} & $47.09$\ci{6.97} & $51.00$\ci{4.58} & $72.93$\ci{4.38} & $87.33$\ci{2.08} & $89.41$\ci{1.75} & $93.00$\ci{1.00} & $95.62$\ci{2.50} \\  [0.4ex]

Ours w/o Double Critic  & $55.67$\ci{3.61} & $60.95$\ci{2.67} & $70.33$\ci{3.06} & $85.64$\ci{3.24} & $94.67$\ci{1.53} & $96.57$\ci{1.42} & $94.33$\ci{3.06} & $95.62$\ci{2.50} \\  [0.4ex]

\textbf{\beamdojo}  & $\mathbf{91.67}$\ci{1.33} & $\mathbf{94.26}$\ci{2.08} & $\mathbf{94.33}$\ci{1.53} & $\mathbf{95.15}$\ci{1.82} & $\mathbf{97.67}$\ci{2.08} & $\mathbf{98.54}$\ci{1.43} & $94.33$\ci{1.15} & $97.00$\ci{1.30} \\

\bottomrule[1.0pt]
\end{tabular}
\end{center}
\end{table*}

We compare our proposed framework \beamdojo, which integrates two-stage RL training and a double critic, with the following baselines:

\begin{enumerate}[label=BL~\arabic*), leftmargin=3em]
    \item \textbf{PIM}~\cite{long2024learninghumanoid}: This one-stage method is designed for humanoid locomotion tasks, such as walking up stairs and traversing uneven terrains. We additionally add our foothold reward $r_\text{foothold}$ to encourage the humanoid to step accurately on the foothold areas.
    \label{baseline:pim}
    
    \item \textbf{Naive}: This method neither include the two-stage RL nor the double critic. The only addition is the foothold reward. This is an naive implementation to solve this task.
    \label{baseline:naive}
    
    \item \textbf{Ours w/o Soft Dyn}: This is an ablation which removing the first stage of training with soft terrain dynamics constraints.
    \label{baseline:no-stage-1}
    
    \item \textbf{Ours w/o Double Critic}: This is an ablation which uses a single critic to handle both locomotion rewards and foothold reward, instead of using a double critic. This follows the traditional design in most locomotion tasks.
    \label{baseline:no-double-critic}
\end{enumerate}

The training and simulation environments are implemented in IsaacGym~\cite{makoviychuk2021isaac}. To ensure fairness, we adapt all methods to two stages. For stage 1, we train the humanoid on the \textit{Stones Everywhere} with curriculum learning. In this stage, our method and \ref{baseline:no-double-critic} use soft terrain dynamics constraints, while all other baselines use hard terrain dynamics constraints. For stage 2, we conduct fine-tuning on the \textit{Stepping Stones} and \textit{Balancing Beams} terrains with curriculum learning.

For evaluation, we test all methods on the \textit{Stepping Stones}, \textit{Balancing Beams}, \textit{Stepping Beams} and \textit{Gaps} terrains. We evaluate performance using three metrics:
\begin{itemize}
    \item \textbf{Success Rate} $R_\mathrm{succ}$: The percentage of successful attempts to cross the entire terrain.
    \item \textbf{Traverse Rate} $R_\mathrm{trav}$: The ratio of the distance traveled before falling to the total terrain length ($8$ m).
    \item \textbf{Foothold Error} $E_\mathrm{foot}$: The average proportion of foot samples landing outside the intended foothold areas.
\end{itemize}

\subsection{Simulation Experiments}

\subsubsection{Quantitative results}

\begin{figure}[t]
    \centering
    \includegraphics[width=\linewidth]{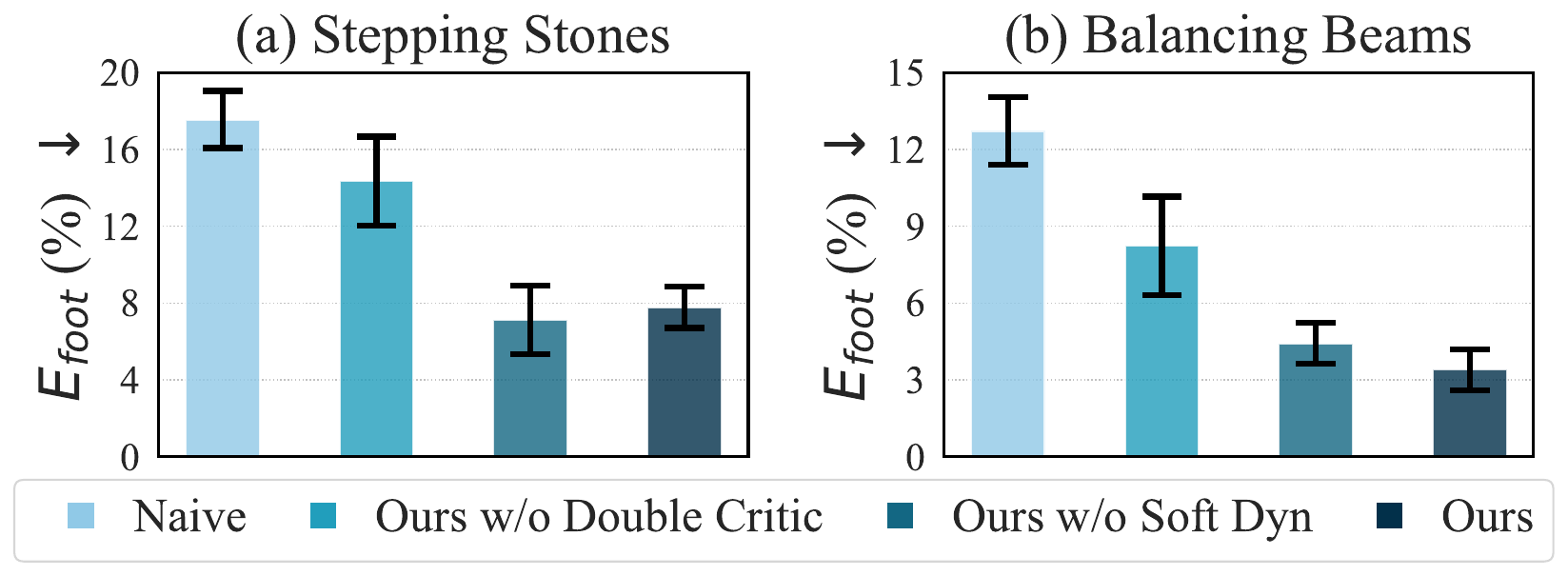}
    \caption{\textbf{Foothold Error Comparison.} The foothold error benchmarks of all methods are evaluated in (a) stepping stones and (b) balancing beams, both tested under medium terrain difficulty.}
    \label{fig:foothold_error}
\end{figure}

We report the success rate ($R_\mathrm{succ}$) and traverse rate ($R_\mathrm{trav}$) for four terrains at medium and hard difficulty levels (terrain level 6 and level 8, respectively) in Table~\ref{tab:all-results}. For each setting, the mean and standard deviation are calculated over three policies trained with different random seeds, each evaluated across 100 random episodes. Our key observations are as follows: 

\begin{itemize}
    \item Leveraging the efficient two-stage RL framework and the double critic, \beamdojo consistently outperforms single-stage approaches and ablation designs, achieving high success rates and low foothold errors across all challenging terrains. Notably, the naive implementation struggles significantly and is almost incapable of traversing stepping stones and balancing beams at hard difficulty levels.
    \item Existing humanoid controllers~\cite{long2024learninghumanoid} face difficulties when adapting to risky terrains with fine-grained footholds, primarily due to sparse foothold rewards and low learning efficiency.
    \item Despite the our method not being explicitly trained on \textit{Stepping Beams} and \textit{Gaps}, it demonstrates impressive zero-shot generalization capabilities on these terrains.
\end{itemize}

\subsubsection{Detailed Ablation Analysis}

\begin{figure}[t]
    \centering
    \includegraphics[width=\linewidth]{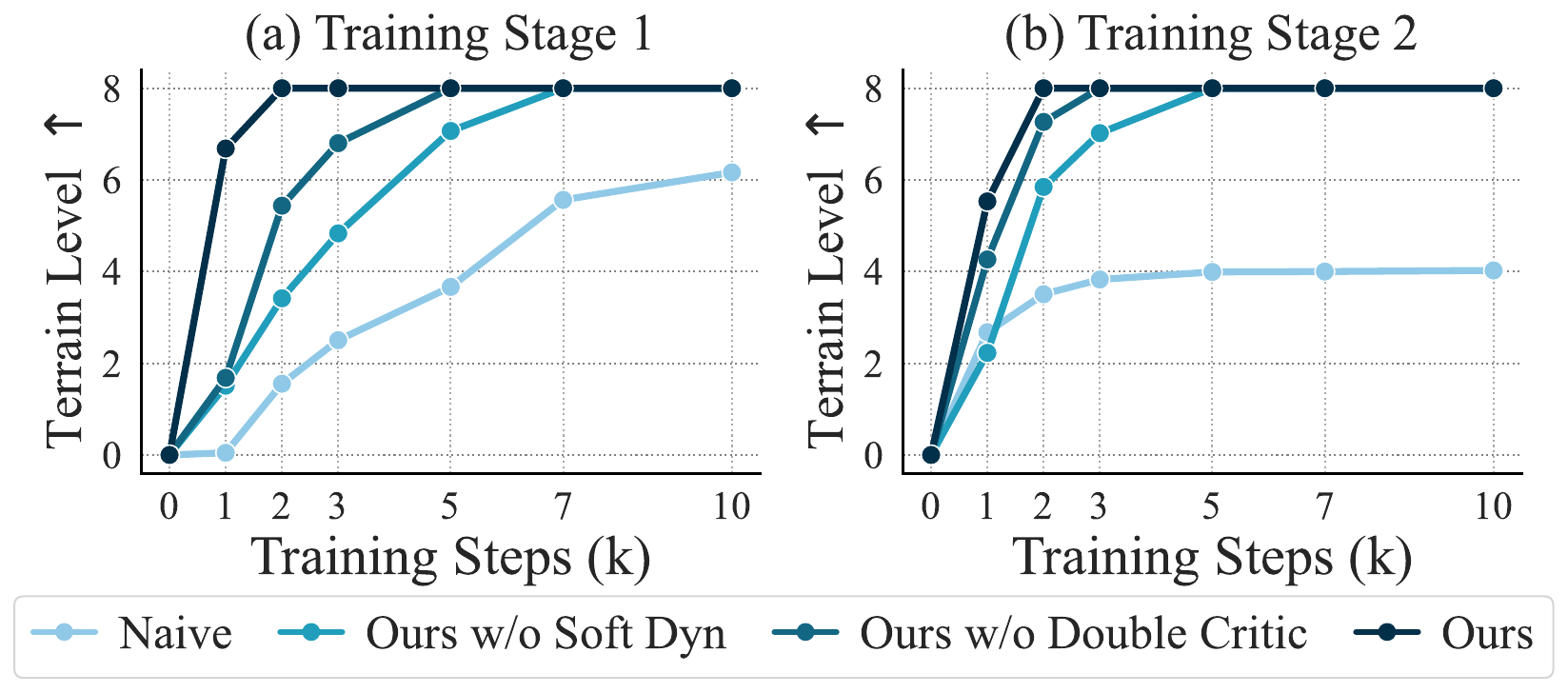}
    \caption{\textbf{Learning Efficiency.} The learning curves show the maximum terrain levels achieved in two training stages of all methods. Faster attainment of terrain level 8 indicates more efficient learning.}
    \label{fig:learning_efficiency}
\end{figure}

We conduct additional ablation studies by comparing \beamdojo with \ref{baseline:naive}, \ref{baseline:no-stage-1}, and \ref{baseline:no-double-critic}.

\textbf{Foot Placement Accuracy:} As shown in Fig.~\ref{fig:foothold_error}, \beamdojo achieves highly accurate foot placement with low foothold error values, largely due to the contribution of the double critic. In comparison, the naive implementation shows higher error rates, with a substantial proportion of foot placements landing outside the safe foothold areas. This demonstrates the precision and effectiveness of our method in challenging terrains.

\textbf{Learning Efficiency:} Although we train for 10,000 iterations in both stages to ensure convergence across all designs, \beamdojo converges significantly faster, as shown in Fig.~\ref{fig:learning_efficiency}. Both the two-stage training setup and the double critic improve learning efficiency, with the two-stage setup contributing the most. In contrast, the naive implementation struggles to reach higher terrain levels in both stages.

The advantage of two-stage learning lies in its ability to allow the agent to continuously attempt foot placements, even in the presence of missteps, making it easier to accumulate a substantial number of successful foot placement samples. Meanwhile, the double-critic setup separates the foothold reward from the locomotion rewards, ensuring that its updates remain unaffected by the noise of unstable locomotion signals, particularly in the early training phase. Both strategies play a crucial role in enhancing learning efficiency.

\begin{table}[t]
    \centering
    \caption{\textbf{Gait Regularization.} We conduct experiments on stepping stones and evaluate three representative gait regularization reward metrics: smoothness, feet air time, and feet clearance. Detailed definitions of the reward functions can be found in Table~\ref{tab:reward}.}
    \begin{tabular}{lcc}
    \toprule[1.0pt]
     \textbf{Designs} & \textbf{Smoothness} ($\downarrow$) & \textbf{Feet Air Time} ($\uparrow$) \\
    \midrule[0.8pt]
    Naive & {$1.7591$\ci{0.1316}} & {$-0.0319$\ci{0.0028}} \\ [0.2ex]
    Ours w/o Soft Dyn & {$0.9633$\ci{0.0526}} & {$\mathbf{-0.0169}$\ci{0.0014}} \\ [0.2ex]
    Ours w/o Double Critic& {$1.2705$\ci{0.1168}} & {$-0.0229$\ci{0.0033}} \\ [0.2ex]
    \textbf{\beamdojo} & {$\mathbf{0.7603}$\ci{0.0315}} & {$-0.0182$\ci{0.0027}} \\ [0.2ex]
    \bottomrule[1.0pt]
    \end{tabular}
    \label{tab:gait_regu}
\end{table}

\begin{figure}[t]
    \centering
    \includegraphics[width=0.95\linewidth]{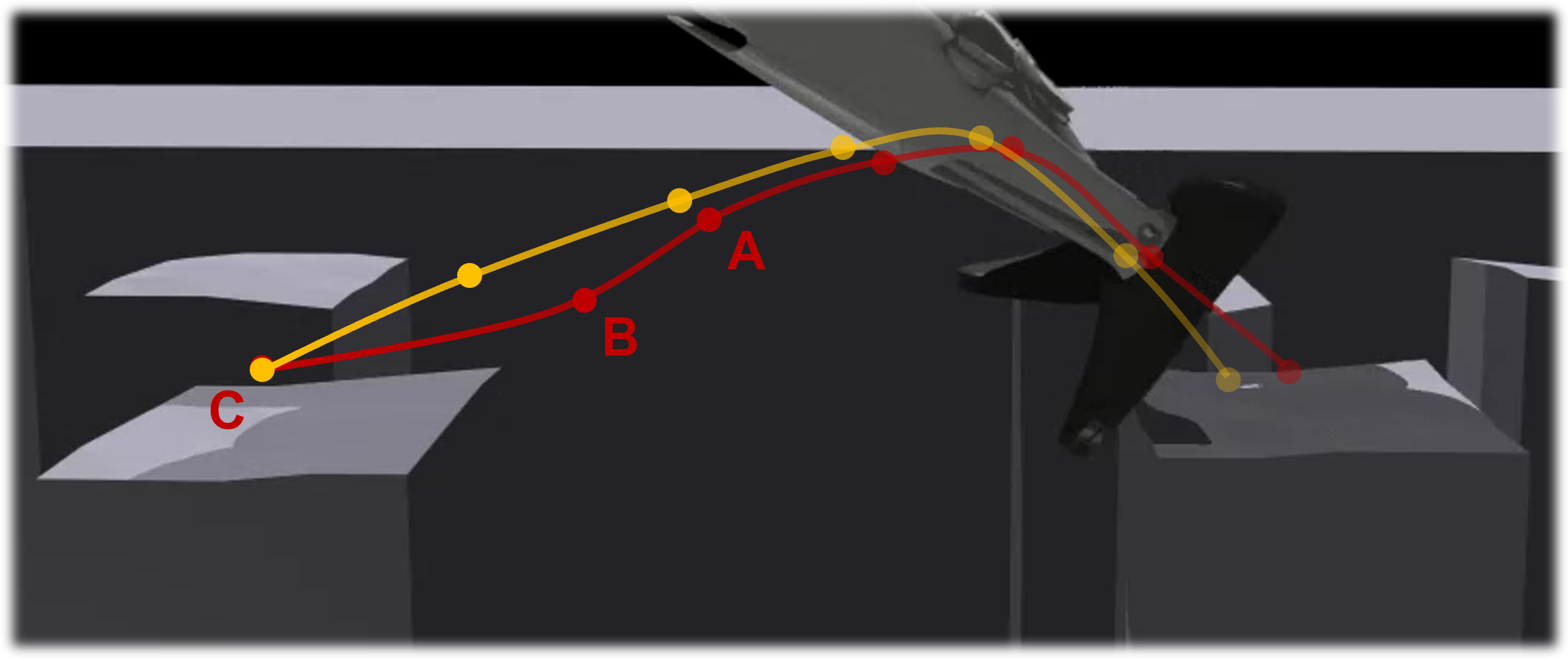}
    \caption{\textbf{Foot Placement Planning Visualization.} We illustrate two trajectories for the foot placement process: the yellow line represents \beamdojo, while the red line corresponds to \textit{Ours w/o Double Critic}. Points along the trajectories are marked at equal time intervals. From A to C, the method without the double critic exhibits significant adjustments only when approaching the target stone (at point B).}
    \label{fig:footplanning}
\end{figure}

\begin{figure*}[!t]
  \begin{minipage}[t]{0.245\linewidth}
    \centering
    \includegraphics[width=4.2cm,height=3.5cm]{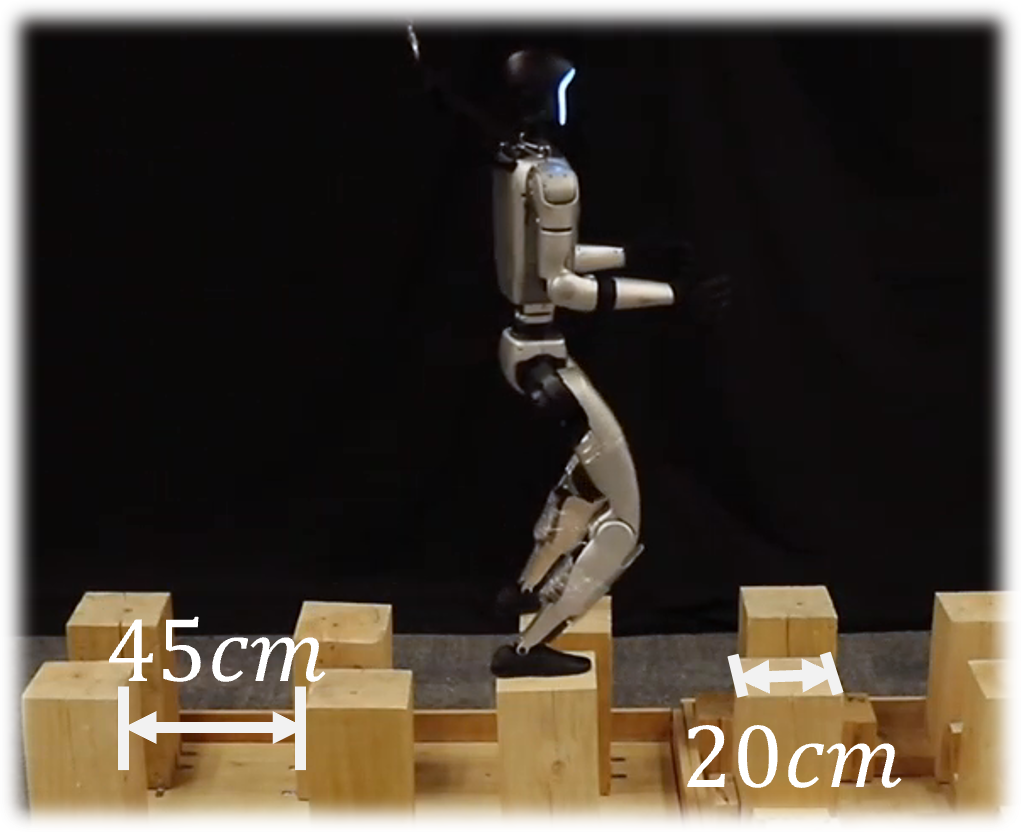}
    \vspace{0.2cm}\\
    \resizebox{1.0\linewidth}{!}{
\begin{tabular}{>{\columncolor[gray]{0.9}}p{2.6cm}cc}
\toprule[1.0pt]
\textit{\textbf{Stepping Stones}} & $R_{\mathrm{succ}}$& $R_\mathrm{trav}$ \\
\midrule[0.8pt]
\rowcolor[gray]{1.0}
Ours w/o HR & $1/5$ & $38.20\%$ \\
\rowcolor[gray]{1.0}
\textbf{\beamdojo} & $4/5$ & $92.18\%$ \\
\bottomrule[1.0pt]
\end{tabular}
}
  \end{minipage}
  \hfill 
  \begin{minipage}[t]{0.245\linewidth}
    \centering
    \includegraphics[width=4.2cm,height=3.5cm]{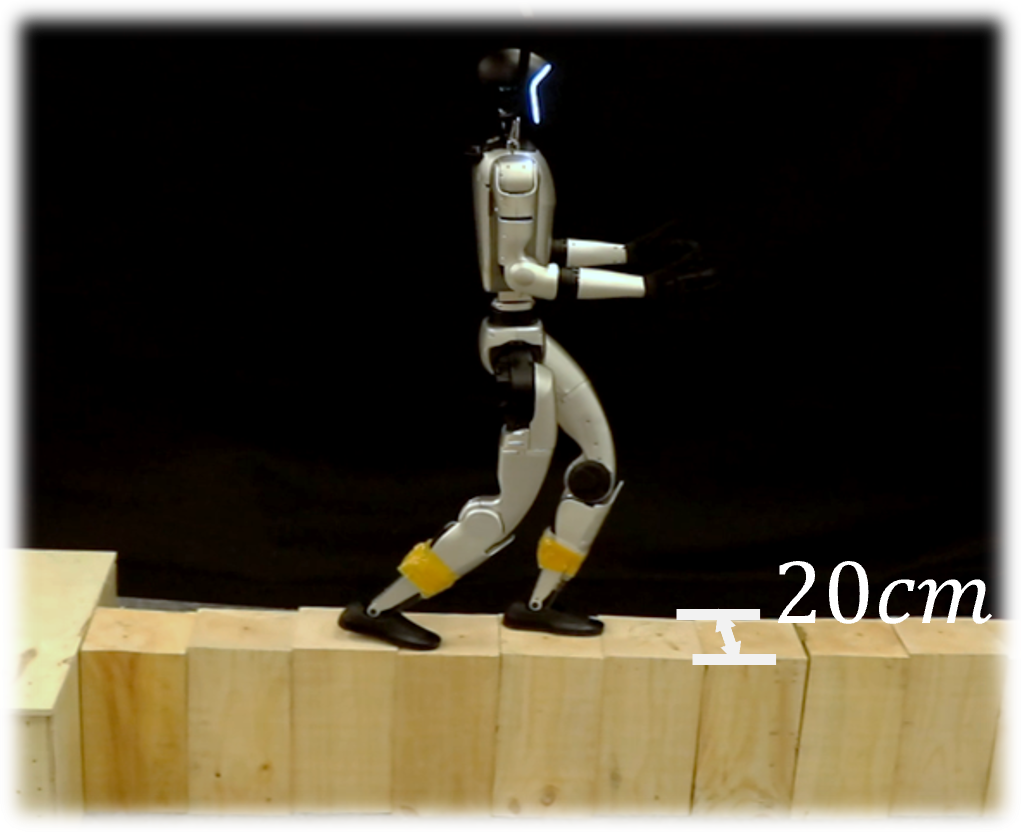}
    \vspace{0.2cm}\\
    \resizebox{1.0\linewidth}{!}{
\begin{tabular}{>{\columncolor[gray]{0.9}}p{2.6cm}cc}
\toprule[1.0pt]
\textit{\textbf{Balancing Beams}} & $R_{\mathrm{succ}}$& $R_\mathrm{trav}$ \\
\midrule[0.8pt]
\rowcolor[gray]{1.0}
Ours w/o HR & $0/5$ & $12.37\%$ \\
\rowcolor[gray]{1.0}
\textbf{\beamdojo} & $4/5$ & $88.16\%$ \\
\bottomrule[1.0pt]
\end{tabular}
}
  \end{minipage}
  \hfill
  \begin{minipage}[t]{0.245\linewidth}
    \centering
    \includegraphics[width=4.2cm,height=3.5cm]{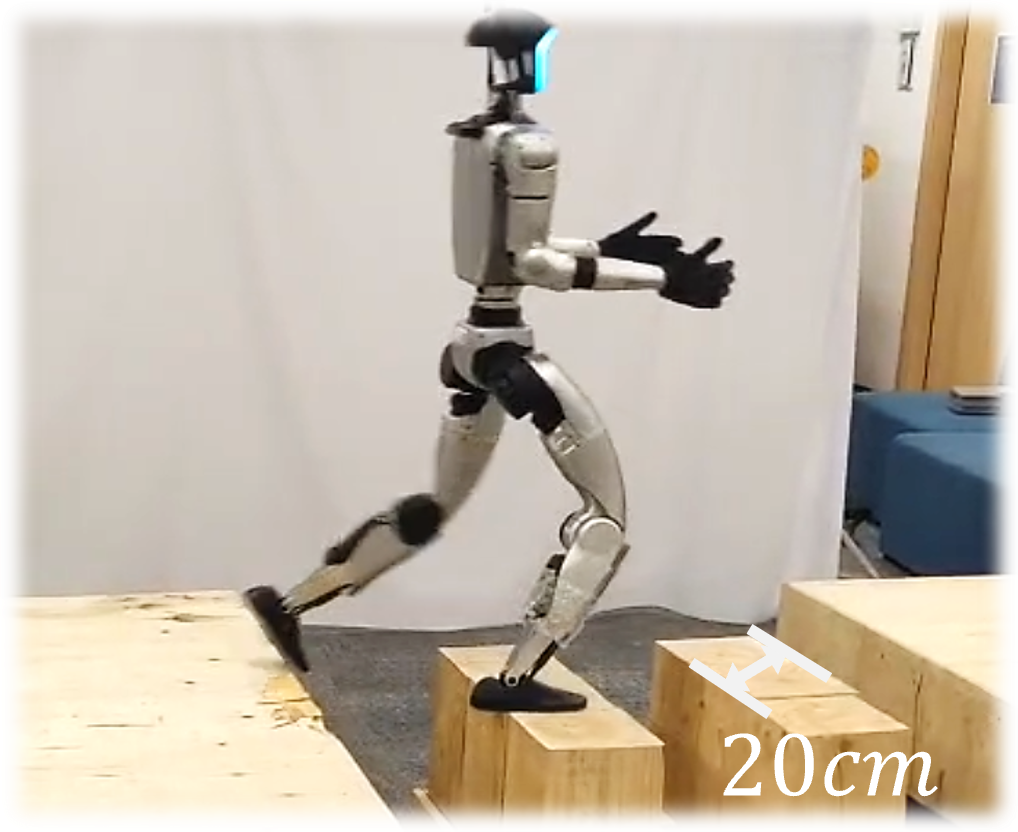}
    \vspace{0.2cm}\\
    \resizebox{1.0\linewidth}{!}{
\begin{tabular}{>{\columncolor[gray]{0.9}}p{2.6cm}cc}
\toprule[1.0pt]
\textit{\textbf{Stepping Beams}} & $R_{\mathrm{succ}}$& $R_\mathrm{trav}$ \\
\midrule[0.8pt]
\rowcolor[gray]{1.0}
Ours w/o HR & $1/5$ & $30.00\%$ \\
\rowcolor[gray]{1.0}
\textbf{\beamdojo} & $3/5$ & $70.00\%$ \\
\bottomrule[1.0pt]
\end{tabular}
}
  \end{minipage}%
  \hfill
  \begin{minipage}[t]{0.245\linewidth}
    \centering
    \includegraphics[width=4.2cm,height=3.5cm]{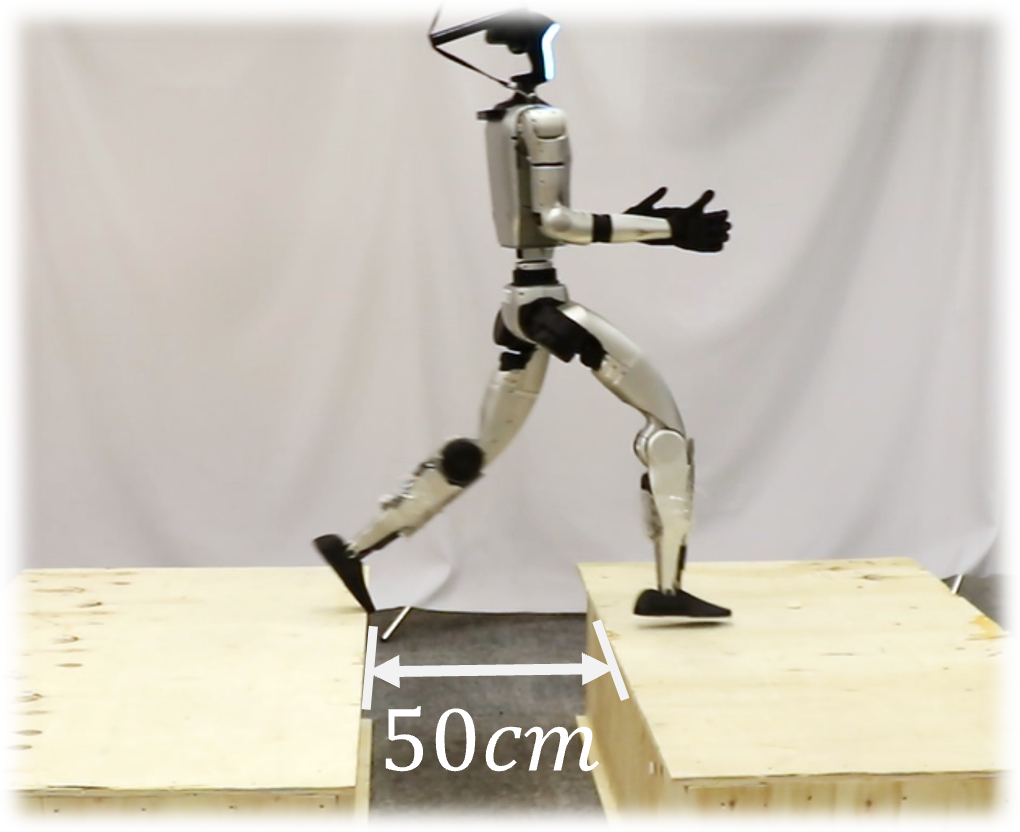}
    \vspace{0.2cm}\\
    \resizebox{1.0\linewidth}{!}{
\begin{tabular}{>{\columncolor[gray]{0.9}}p{2.6cm}cc}
\toprule[1.0pt]
\textit{\textbf{Gaps}} & $R_{\mathrm{succ}}$& $R_\mathrm{trav}$ \\
\midrule[0.8pt]
\rowcolor[gray]{1.0}
Ours w/o HR & $3/5$ & $60.00\%$ \\
\rowcolor[gray]{1.0}
\textbf{\beamdojo} & $5/5$ & $100.00\%$ \\
\bottomrule[1.0pt]
\end{tabular}
}
  \end{minipage}

\caption{\textbf{Real-world Experiments.} We build terrains in the real world similar to those in simulation. (a) \textbf{Stepping Stones:} stones with a size of 20 cm, a maximum distance of 45 cm between stones, and a sparsity level of 72.5\%. (b) \textbf{Balancing Beams:} beams with a width of 20 cm. (c) \textbf{Stepping Beams:} beams with a size of 20 cm, a maximum distance of 45 cm between beams, and a sparsity level of 66.6\%. (d) \textbf{Gaps:} gaps with a maximum distance of 50 cm.}
\label{fig:real_results}
\end{figure*}

\begin{figure*}[t]
    \centering
    \includegraphics[width=\linewidth]{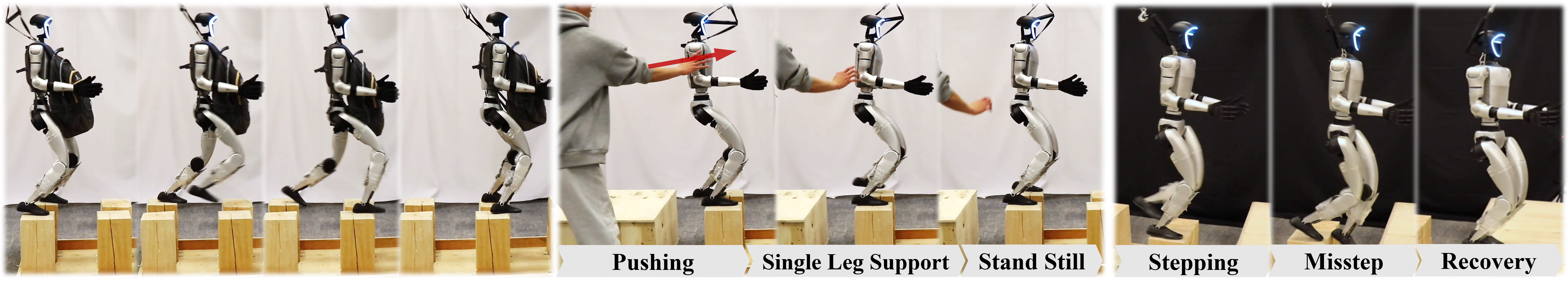}
    \caption{\textbf{Robustness Test.} We evaluate the robustness of the humanoid robot in real-world scenarios with: (a) heavy payload, (b) external forces, and (c) recovering from missteps.}
    \label{fig:robustness}
\end{figure*}

\textbf{Gait Regularization:} The combination of small-scale gait regularization rewards with sparse foothold reward can hinder gait performance, as shown in Table~\ref{tab:gait_regu}, where the naive design and the ablation without the double critic exhibit poor performance in both smoothness and feet air time. In contrast, our method and the ablation with double critic demonstrates superior motion smoothness and improved feet clearance. This improvement arises because, in the double-critic framework, the advantage estimates for the dense and sparse reward groups are normalized independently, preventing the sparse rewards from introducing noise that could disrupt the learning of regularization rewards.

\textbf{Foot Placement Planning:} As illustrated in Fig.~\ref{fig:footplanning}, we observe that the double critic also benefits foot placement planning of the entire sub-process of foot lifting and landing. Our method, \beamdojo, enables smoother planning, allowing the foot to precisely reach the next foothold. In comparison, the baseline excluding double critic demonstrates reactive stepping, where adjustments are largely made when the foot is close to the target stone. This behavior indicates that the double critic, by learning the sparse foothold reward separately, helps the policy adjust its motion over a longer horizon.

\subsection{Real-world Experiments}

\subsubsection{Result}

As demonstrated in Fig.\ref{fig:teaser}, our framework achieves zero-shot transfer, successfully generalizing to real-world dynamics. To showcase the effect of height map domain randomization (introduced in Section~\ref{sec:sim2real}) in sim-to-real transfer, we compare our proposed method with an ablation that excludes height map randomization (denoted as ``ours w/o HR"). We conduct five trials on each terrain and report the success and traversal rates in Fig.~\ref{fig:real_results}, with following conclusions:
\begin{itemize}
    \item \beamdojo achieves a high success rate in real-world deployments, demonstrating excellent precise foot placement capabilities. Similar to simulation results, it also exhibits impressive generalization performance on \textit{Stepping Beams} and \textit{Gaps}, even though these terrains were not part of the training set.
    \item The ablation, lacking height map domain randomization, results in a significantly lower success rate, highlighting the importance of this design.
    \item It is also worth mentioning that \beamdojo enables backward movement in risky terrains, as shown in Fig.~\ref{fig:teaser}(b). This advantage is achieved by leveraging LiDAR to its full potential, whereas a single depth camera cannot handle such scenarios.
\end{itemize}

\subsubsection{Agility Test}

\begin{table}[t]
    \centering
    \caption{\textbf{Agility Test.} We evaluate the agility of the humanoid robot on stepping stones with a total length of 2.8m.}
    \begin{tabular}{cccc}
    \toprule[1.0pt]
    \textbf{$\mathbf{v}_x^c$} (m/s) & \textbf{Time Cost} (s) & \textbf{Average Speed} (m/s) & \textbf{Error Rate} (\%, $\downarrow$) \\
    \midrule[0.8pt]
    $0.5$ & {$6.33$\ci{0.15}} & {$0.45$\ci{0.05}} & {$10.67$\ci{4.54}} \\ [0.2ex]
    $0.75$ & {$4.33$\ci{0.29}} & {$0.65$\ci{0.05}} & {$13.53$\ci{6.52}} \\ [0.2ex]
    $1.0$ & {$3.17$\ci{0.58}} & {$0.88$\ci{0.04}} & {$11.83$\ci{8.08}} \\ [0.2ex]
    $1.25$ & {$2.91$\ci{0.63}} & {$0.96$\ci{0.03}} & {$22.74$\ci{5.32}} \\ [0.2ex]
    $1.5$ & {$2.69$\ci{0.42}} & {$1.04$\ci{0.05}} & {$30.68$\ci{6.17}} \\ [0.2ex]
    \bottomrule[1.0pt]
    \end{tabular}
    \label{tab:agile_test}
\end{table}

To assess the agility of our method, we provide the humanoid robot with five commanded longitudinal velocities, $\mathbf{v}_x^c$: 0.5, 0.75, 1.0, 1.25, and 1.5\,m/s, and check the tracking ability. Each test was conducted over three trials, and the results are reported in Table~\ref{tab:agile_test}. The results show minimal tracking error up to the highest training command velocity of 1.0\,m/s, where the robot achieves an average speed of 0.88\,m/s, demonstrating the agility of our policy. However, performance degrades significantly beyond 1.25\,m/s, as maintaining such speeds becomes increasingly difficult on these highly challenging terrains.

\subsubsection{Robustness Test}

To evaluate the robustness of our precise foothold controller, we conducted the following experiments on real-world experiment terrains:
\begin{itemize}
    \item \textbf{Heavy Payload}:As shown in Fig.~\ref{fig:robustness}(a), the robot carried a 10 kg payload—approximately 1.5 times the weight of its torso—causing a significant shift in its center of mass. Despite this challenge, the robot effectively maintained agile locomotion and precise foot placements, demonstrating its robustness under increased payload conditions.
    \item \textbf{External Force}: As shown in Fig.~\ref{fig:robustness}(b), the robot was subjected to external forces from various directions. Starting from a stationary pose, the robot experienced external pushes, transitioned to single-leg support, and finally recovered to a stable standing position with two-leg support.
    \item \textbf{Misstep Recovery}: As shown in Fig.~\ref{fig:robustness}(c), the robot traverse terrain without prior scanning of terrain dynamics. Due to occlusions, the robot lacked information about the terrain underfoot, causing initial missteps. Nevertheless, it demonstrated robust recovery capabilities.
\end{itemize}

\subsection{Extensive Studies and Analysis}

\subsubsection{Design of Foothold Reward}

\begin{table}[t]
    \centering
    \setlength{\tabcolsep}{12pt}
    \caption{\textbf{Comparison of Different Foothold Reward Designs.} The success rate and foothold error for each foothold reward design are evaluated on stepping stones with medium terrain difficulty.}
    \begin{tabular}{lcc}
    \toprule[1.0pt]
    Designs & $R_\mathrm{succ}$ ($\%, \uparrow$) & $E_\mathrm{foot}$ ($\%, \downarrow$) \\
    \midrule[0.8pt]
    foothold-$30\%$ & $93.67$\ci{1.96} & $11.43$\ci{0.81} \\ [0.2ex]
    foothold-$50\%$ & $92.71$\ci{1.06} & $10.78$\ci{1.94} \\ [0.2ex]
    foothold-$70\%$ & $91.94$\ci{2.08} & $14.35$\ci{2.61} \\ [0.2ex]
    \textbf{\beamdojo} & $\mathbf{95.67}$\ci{1.53}  & $\mathbf{7.79}$\ci{1.33} \\
    
    \bottomrule[1.0pt]
    \end{tabular}
    \label{tab:foothold_reward}
\end{table}

As discussed in Section~\ref{sec:method_footholdreward}, our sampling-based foothold reward is proportional to the number of safe points, making it a relatively continuous reward: the larger the overlap between the foot placement and the safe footholds, the higher the reward the agent receives. We compare this approach with other binary and coarse reward designs: when $p\%$ of the sampled points fall outside the safe area, a full penalty is applied; otherwise, no penalty is given. This can be defined as:
\begin{equation}
    r_{\text{foothold}-p\%} = -\sum_{i=1}^2 \mathbb{C}_i \cdot \mathds{1} \left\{ \left( \sum_{j=1}^n \mathds{1} \{ d_{ij} < \epsilon \} \right) \ge p\% \cdot  n\right\}.
\end{equation}

We compare our continuous foothold reward design with three variants of the coarse-grained approach, where $p=30, 50$, and $70$ (denoted as foothold-$30\%$, foothold-$50\%$, and foothold-$70\%$ respectively). The success rate $R_\mathrm{succ}$ and the foothold error $E_\mathrm{foot}$ on stepping stones are reported in Table~\ref{tab:foothold_reward}.

It is clear that our fine-grained design enables the robot to make more accurate foot placements compared to the other designs, as this continuous approach gradually encourages maximizing the overlap. Among the coarse-grained approaches, foothold-$50\%$ performs better than foothold-$30\%$ and foothold-$70\%$, as a $30\%$ threshold is too strict to learn effectively, while $70\%$ is overly loose.

\subsubsection{Design of Curriculum}

\begin{table}[t]
    \centering
    \setlength{\tabcolsep}{8pt}
    \caption{\textbf{Comparison of Different Curriculum Designs.} The success rate and traverse rate for each curriculum design are evaluated on stepping stones with medium and hard terrain difficulty respectively.}
    \begin{tabular}{lcccc}
    \toprule[1.0pt]
    \multirow{2}{*}{Designs} & \multicolumn{2}{c}{Medium Difficulty} & \multicolumn{2}{c}{Hard Difficulty} \\
    \cmidrule[\heavyrulewidth](lr){2-3}
    \cmidrule[\heavyrulewidth](lr){4-5} 
     & $R_\mathrm{succ}$ & $R_\mathrm{trav}$ & $R_\mathrm{succ}$ & $R_\mathrm{trav}$ \\
    \midrule[0.8pt]
    w/o curriculum-medium & $88.33$ & $90.76$ & $2.00$ &  $18.36$ \\ [0.2ex]
    w/o curriculum-hard & $40.00$ & $52.49$ & $23.67$ & $39.94$ \\ [0.2ex]
    \textbf{\beamdojo} & $\mathbf{95.67}$ & $\mathbf{96.11}$ & $\mathbf{82.33}$ & $\mathbf{86.87}$\\
    
    \bottomrule[1.0pt]
    \end{tabular}
    \label{tab:curriculum}
\end{table}

To validate the effectiveness of the terrain curriculum introduced in Section~\ref{sec:method_curriculum}, we introduce an ablation study without curriculum learning. In this design, we train using only medium and hard terrain difficulties in both stages (denoted as ``w/o curriculum-medium'' and ``w/o curriculum-hard''). Similarly, we report the $R_\mathrm{succ}$ and $R_\mathrm{trav}$ for both ablation methods, along with our method, on stepping stones terrain at two different difficulty levels in Table~\ref{tab:curriculum}. The results show that incorporating curriculum learning significantly improves both performance and generalization across terrains of varying difficulty. In contrast, without curriculum learning, the model struggles significantly with challenging terrain when learning from scratch  (``ours w/o curriculum-hard''), and also faces difficulties on other terrain types, severely limiting its generalization ability (``ours w/o curriculum-medium'').

\subsubsection{Design of Commands}

As mentioned in Section~\ref{sec:method_curriculum}, in the second stage, no heading command is applied, and the robot is required to learn to consistently face forward through terrain dynamics. We compare this approach with one that includes a heading command (denoted as ``ours w/ heading command''), where deviation from the forward direction results in a corrective yaw command based on the current directional error. In the deployment, we use the LiDAR odometry module to update the heading command in real time, based on the difference between the current orientation and the initial forward direction.

We conduct five trials on the stepping stones terrain in the real world, comparing our proposed method with the ``ours w/ heading command'' design. The success rates are $4/5$ and $1/5$, respectively. The poor performance of the heading command design is primarily due to two factors: (1). In the simulation, the model overfits the angular velocity of the heading command, making it difficult to handle noisy real-world odometry data; (2). In the real world, precise manual calibration of the initial position is required to determine the correct forward direction, making the heading command approach less robust. In contrast, \beamdojo, without heading correction, proves to be more effective.

\subsubsection{Generalization to Non-Flat Terrains}

\begin{figure}[t]
    \centering
    \includegraphics[width=\linewidth]{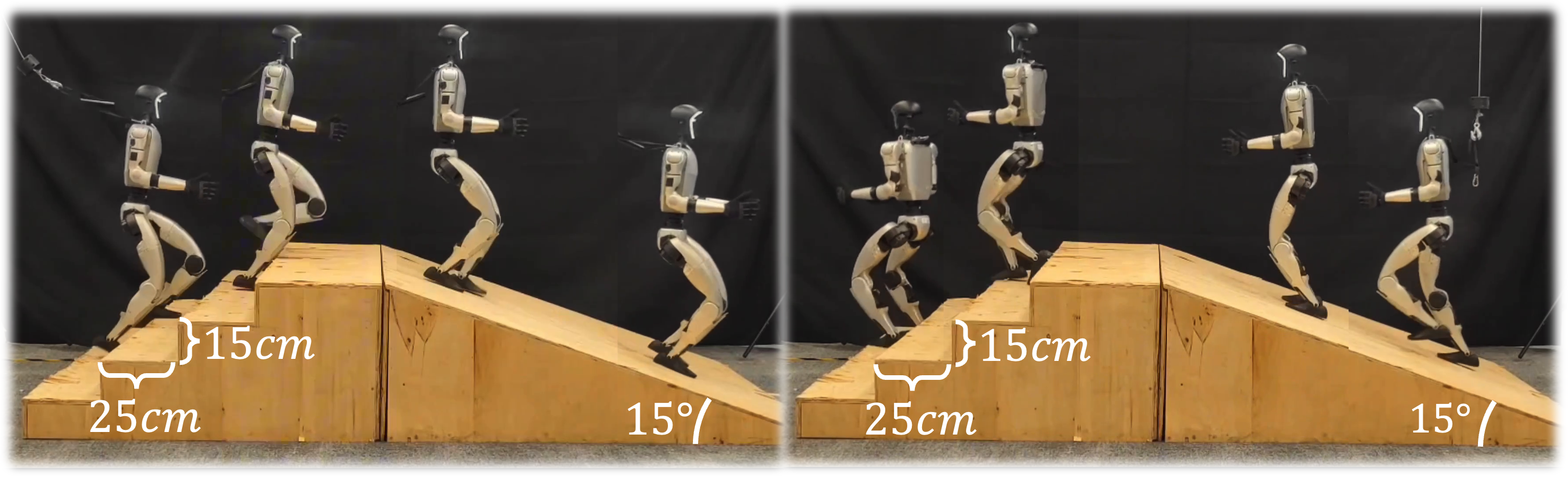}
    \caption{\textbf{Generalization Test on Non-Flat Terrains.} We conduct real-world experiments on (a) \textbf{stairs} with a width of 25\,cm and a height of 15\,cm, and (b) \textbf{slopes} with a 15-degree incline.}
    \label{fig:stair_slope}
\end{figure}

We observe that \beamdojo also generalizes well to non-flat terrains such as stairs and slopes. The main adaptation involves calculating the base height reward relative to the foot height rather than the ground height on these uneven surfaces. Additionally, Stage,1 pre-training becomes unnecessary for stairs and slopes, as the footholds are no longer sparse. We validate our approach through hardware experiments on stairs and slopes, as shown in Fig.~\ref{fig:stair_slope}. The results demonstrate that \beamdojo enables the robot to successfully traverse stairs and slopes with success rates of 8/10 and 10/10, respectively.

\subsubsection{Failure Cases}

\begin{figure}[t]
    \centering
    \includegraphics[width=\linewidth]{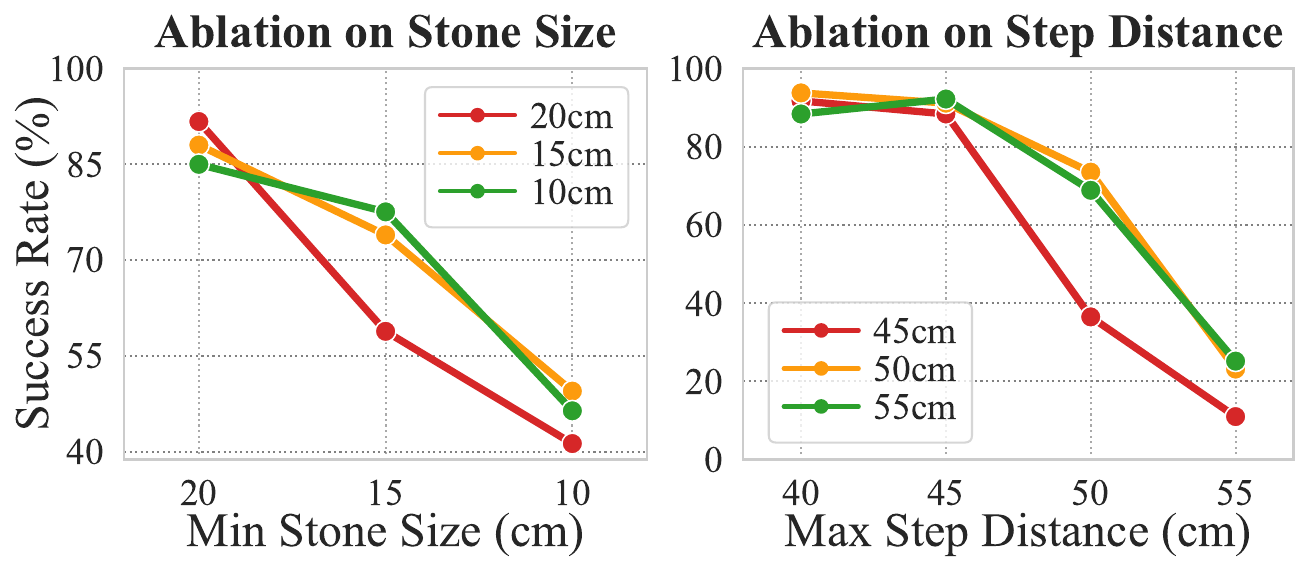}
    \caption{\textbf{Failure Case Analysis.} We evaluate the success rate on varying (a) stone sizes, and (b) step distances.}
    \label{fig:failure_case}
\end{figure}

To investigate the framework's performance limitations, we evaluate its performance across varying stone sizes and step distances, as shown in Fig.~\ref{fig:failure_case}. We compare policies trained with different terrain parameters, with the smallest stone size (20\,cm, 15\,cm and 10\,cm) or largest step distance (45\,cm, 50\,cm and 55\,cm). The results indicate that while tougher training enhances adaptability, performance still drops sharply on 10\,cm stones (approximately half the foot length) and 55\,cm steps (roughly equal to the leg length), even under the most challenging training settings. In these cases, the difficulty shifts toward maintaining balance on very small footholds and executing larger strides—challenges that the current reward function does not adequately address.

\subsubsection{Limitations}

On the one hand, the performance of our method is significantly constrained by the limitations of the perception module. Inaccuracies in LiDAR odometry, along with issues such as jitter and map drift, present considerable challenges for real-world deployment. Furthermore, when processing LiDAR data, the trade-off between the confidence in noisy measurements and the dynamic changes in terrain—such as the jitter of stones, which is difficult to simulate in the simulation—makes it challenging to effectively handle sudden disturbances or variations. As a result, the system struggles to quickly and flexibly adapt to unexpected changes in the environment.

On the other hand, our method has yet to fully leverage the information provided by the elevation map, and has not adequately addressed the challenges of terrains with significant foothold height variations. In future work, we aim to develop a more generalized controller that enables agile locomotion, extending to a broader range of terrains, including stairs and other complex surfaces that require footstep planning, as well as terrains with significant elevation changes.

\section{Conclusion}

In this paper, we proposed a novel framework, \beamdojo, which enables humanoid robots to traverse with agility and robustness on sparse foothold terrains such as stepping stones and balance beams, and generalize to a wider range of challenging terrains (e.g., gaps, balancing beams). The key conclusions are summarized as follows:

\begin{itemize}
    \item \textbf{Accuracy of Foot Placement:} We introduced a foothold reward for polygonal feet, which is proportional to the contact area between the footstep and the safe foothold region. This continuous reward effectively encourages precise foot placements.
    
    \item \textbf{Training Efficiency and Effectiveness:} By incorporating a two-stage reinforcement learning training process, \beamdojo enables full trial-and-error exploration. Additionally, the double-head critic significantly enhances the learning of sparse foothold rewards, regularizes gait patterns, and facilitates long-distance foot placement planning.
    
    \item \textbf{Agility and Robustness in the Real World:} Our experiments demonstrate that \beamdojo empowers humanoid robots to exhibit agility and achieve a high success rate in real-world scenarios. The robots maintain stable walking even under substantial external disturbances and the inevitable sway of beams in real world. Notably, by leveraging LiDAR-based mapping, we have achieved stable backward walking, a challenge typically encountered with depth cameras.
\end{itemize}

\section*{Acknowledgments}

This work is funded in part by the National Key R\&D Program of China (2022ZD0160201), and Shanghai Artificial Intelligence Laboratory. We thank the authors of PIM~\cite{long2024learninghumanoid} for their kind help with the deployment of the elevation map. We thank Unitree Robotics for their help with the Unitree G1 humanoid robot. We thank RSS reviewers for their careful and insightful feedback, which helped improve the quality of this work.

\vspace{0.2cm}

\bibliographystyle{plainnat}
\bibliography{references}

\clearpage
\section*{Appendix}

\subsection{Reward Functions}
\label{sec:append_reward}
The reward functions we used during the training are shown in Table~\ref{tab:reward}, which mainly comes from~\cite{agarwal2023legged, fu2021minimizing, kumar2021rma, margolis2023walk, long2024learninghumanoid}. The corresponding symbols and their descriptions are provided in Table~\ref{tab:symbol}.
\vspace{-0.1cm}

\begin{table}[h]
    \centering
    \caption{Reward Functions}
    \begin{tabular}{lll}
    \toprule[1.0pt]
    \textbf{Term} & \textbf{Equation} & \textbf{Weight} \\
    
    \midrule[0.8pt]
    \multicolumn{3}{c}{\textbf{Group 1: Locomotion Reward Group}} \\ [0.2ex]
    xy velocity tracking & $\exp \left\{- {\|\mathbf{v}_{x y}-\mathbf{v}_{x y}^c\|_2^2}/{\sigma}  \right\}$ & $1.0$ \\
    yaw velocity tracking & $\exp \left\{- {\left(\boldsymbol{\omega}_{\text {yaw}}-\boldsymbol{\omega}_{\text {yaw}}^c\right)^2}/{\sigma} \right\}$ & $1.0$ \\
    
    base height & $\left(h - h^{\text {target}}\right)^2$ & $-10.0$ \\ [0.2ex]
    orientation & $\|\mathbf{g}_{x}\|_2^2 + \|\mathbf{g}_{y}\|_2^2$ & $-2.0$ \\ [0.2ex]
    z velocity & $\mathbf{v}_z^2$ & $-2.0$ \\ [0.2ex]
    roll-pitch velocity & $\|\boldsymbol{\omega}_{x y}\|_2^2$ & $-0.05$ \\ [0.2ex]
    action rate & $\|\mathbf{a}_t-\mathbf{a}_{t-1}\|_2^2$ & $-0.01$ \\ [0.2ex]
    smoothness & $\|\mathbf{a}_t-2 \mathbf{a}_{t-1}+\mathbf{a}_{t-2}\|_2^2$ & $-1e-3$ \\ [0.2ex]
    stand still & $\|{\boldsymbol{\theta}}\|_{2}^{2} \cdot \mathds{1}\{ \| \mathbf{v}_{x y}^c \|_2^2 < \epsilon \}$ & $-0.05$  \\ [0.2ex]
    
    joint velocities & $\|\dot{\boldsymbol{\theta}}\|_{2}^{2}$ & $-1e-4$ \\ [0.2ex]
    joint accelerations & $\|\ddot{\boldsymbol{\theta}}\|_2^2$ & $-2.5e-8$ \\ [0.2ex]
    joint position limits & \makecell[l]{
        $\text{ReLU}({\boldsymbol\theta} - {\boldsymbol\theta}_\text{max}) +$ \\
        \quad \quad \quad $\text{ReLU}({\boldsymbol\theta_\text{min}} - {\boldsymbol\theta})$
    } & $-5.0$ \\ [1.5ex]
    joint velocity limits & $\text{ReLU}(|\dot{\boldsymbol\theta}| - |\dot{\boldsymbol\theta}_\text{max}|)$ & $-1e-3$ \\ [0.2ex]
    joint power & ${|\boldsymbol{\tau} \| \dot{\boldsymbol{\theta}}|^{T}}/\left( \|\mathbf{v}\|_2^2 + 0.2 * \|\boldsymbol{\omega}\|_2^2 \right)$ & $-2e-5$ \\ [0.2ex]
    
    feet ground parallel & $\sum_{i=1}^2 \text{Var}(\mathbf{p}_{z, i})$ & $-0.02$ \\ [0.2ex]
    feet distance & $\text{ReLU} \left( |\mathbf{p}_{y, 1} - \mathbf{p}_{y, 2}| - d_\text{min} \right)$ & $0.5$ \\ [0.2ex]
    feet air time & $\sum_{i=1}^2 \left( t_{\text{air}, i} - t_\text{air}^\text{target} \right) \cdot \mathbb{F}_i$ & $1.0$ \\ [0.2ex]
    feet clearance & $\sum_{i=1}^2 \left(\mathbf{p}_{z, i} - \mathbf{p}_{z}^\text{target} \right)^2 \cdot \dot{\mathbf{p}}_{xy, i}$ & $-1.0$ \\ [0.2ex]
    
    \midrule[0.5pt] 
    \multicolumn{3}{c}{\textbf{Group 2: Foothold Reward Group}} \\ [0.2ex]
    foothold & $-\sum_{i=1}^2 \mathbb{C}_i \sum_{j=1}^n \mathds{1} \{ d_{ij} < \epsilon \}$ & $1.0$ \\ [0.2ex]
    \bottomrule[1.0pt]
    \end{tabular}
    \label{tab:reward}
\end{table}
\vspace{-0.35cm}
\begin{table}[h]
    \centering
    \caption{Used Symbols}
    \begin{tabular}{c p{7cm}}
    \toprule[1.0pt]
    \textbf{Symbols} & \textbf{Description} \\
    \midrule[0.8pt]
    
    $\sigma$ & Tracking shape scale, set to $0.25$. \\ [0.2ex]
    $\epsilon$ & Threshold for determining zero-command in stand still reward, set to $0.1$. \\ [0.2ex]
    $\boldsymbol{\tau}$ & Computed joint torques. \\ [0.2ex]
    $h^\text{target}$ & Desired base height relative to the ground, set to $0.725$. \\ [0.2ex]
    ReLU($\cdot$) & Function that clips negative values to zero~\cite{fukushima1969visual}. \\ [0.2ex]
    $\mathbf{p}_i, \dot{\mathbf{p}}_i$ & Spatial position and velocity of all sampled points on the $i$-th foot respectively. \\ [0.2ex]
    $\mathbf{p}_{z}^\text{target}$ & Target foot-lift height, set to $0.1$. \\ [0.2ex]
    $t_{\text{air}, i}$ & Air time of the $i$-th foot. \\ [0.3ex]
    $t_{\text{air}}^\text{target}$ & Desired feet air time, set to $0.5$. \\ [0.3ex]
    $\mathbb{F}_i$ & Indicator specifying whether foot $i$ makes first ground contact. \\ [0.2ex]
    $d_\text{min}$ & Minimum allowable distance between two feet, set to $0.18$. \\
    \bottomrule[1.0pt]
    \end{tabular}
    \label{tab:symbol}
\end{table}
\vspace{-0.1cm}

\subsection{Terrain Curriculum}
\label{sec:append_curriculum}
The training terrains using curriculum comprises \textit{Stones Everywhere}, \textit{Stepping Stones}, and \textit{Balancing Beams}. The \textit{Stones Everywhere} terrain spans an area of $8m\times8m$, while both \textit{Stepping Stones} and \textit{Balancing Beams} are $2m$ in width and $8m$ in length, with single-direction commands. The depth of gaps relative to the ground is set to $1.0m$, and all stones and beams exhibit height variations within $\pm 0.05m$. The depth tolerance threshold, $\epsilon$, is set to $-0.1 m$.

We define terrain difficulty levels ranging from 0 to 8, denoted as $l$. The specific terrain curriculum at each difficulty level are as follows:
\begin{itemize}
    \item \textit{Stones Everywhere}: The stone size is $\max\{0.25, 1.5(1-0.1l)\}$, and the stone distance is $0.05\lceil l/2\rceil$.
    \item \textit{Stepping Stones}: The stone sizes follow the sequence $[0.8, 0.65, 0.5, 0.4, 0.35, 0.3, 0.25, 0.2, 0.2]$, with a maximum stone distance of $0.1 + 0.05l$.
    \item \textit{Balancing Beams}: The stone size is $0.3-0.05\lfloor l/3\rfloor$, with the stone distance in $x$-direction $0.4 - 0.05l$, and in $y$-direction $[0.2, 0.2, 0.2, 0.25, 0.3, 0.35, 0.35, 0.4, 0.2]$. At the highest difficulty level, the terrain forms a single continuous balancing beam.
\end{itemize}

\subsection{Domain Randomization}
\label{sec:append_random}
\begin{table}[h]
    \centering
    \caption{Domain Randomization Setting}
    \begin{tabular}{ll}
    \toprule[1.0pt]
    \textbf{Term} & \textbf{Value}\\
    
    \midrule[0.8pt]
    \multicolumn{2}{c}{\textbf{Observations}} \\ [0.3ex]
    angular velocity noise & $\mathcal{U}(-0.5, 0.5)$ rad/s \\ 
    joint position noise & $\mathcal{U}(-0.05, 0.05)$ rad/s \\ 
    joint velocity noise & $\mathcal{U}(-2.0, 2.0)$ rad/s \\ 
    projected gravity noise & $\mathcal{U}(-0.05, 0.05)$ rad/s \\ 
    
    \midrule[0.5pt] 
    \multicolumn{2}{c}{\textbf{Humanoid Physical Properties}} \\ [0.3ex]
    actuator offset & $\mathcal{U}(-0.05, 0.05)$ rad \\ 
    motor strength noise & $\mathcal{U}(0.9, 1.1)$ \\ 
    payload mass & $\mathcal{U}(-2.0, 2.0)$ kg \\ 
    center of mass displacement & $\mathcal{U}(-0.05, 0.05)$ m \\ 
    Kp, Kd noise factor & $\mathcal{U}(0.85, 1.15)$ \\ 

    \midrule[0.5pt] 
    \multicolumn{2}{c}{\textbf{Terrain Dynamics}} \\ [0.3ex]
    friction factor & $\mathcal{U}(0.4, 1.0)$ \\ 
    restitution factor & $\mathcal{U}(0.0, 1.0)$ \\ 
    terrain height noise & $\mathcal{U}(-0.02, 0.02)$ m \\ 

    \midrule[0.5pt] 
    \multicolumn{2}{c}{\textbf{Elevation Map}} \\ [0.3ex]
    vertical offset & $\mathcal{U}(-0.03, 0.03)$ m \\ 
    vertical noise & $\mathcal{U}(-0.03, 0.03)$ m \\ 
    map roll, pitch rotation noise & $\mathcal{U}(-0.03, 0.03)$ m \\ 
    map yaw rotation noise & $\mathcal{U}(-0.2, 0.2)$ rad \\ 
    foothold extension probability & $0.6$ \\ 
    map repeat probability & $0.2$ \\ 
    
    \bottomrule[1.0pt]
    \end{tabular}
    \label{tab:domain_random}
\end{table}
\vspace{-0.15cm}

\subsection{Hyperparameters}
\vspace{-0.15cm}
\begin{table}[h]
    \centering
    \caption{Hyperparameters}
    \begin{tabular}{ll}
    \toprule[1.0pt]
    \textbf{Hyperparameter} & \textbf{Value} \\

    \midrule[0.8pt] 
    \multicolumn{2}{c}{\textbf{General}} \\
    num of robots & 4096 \\
    num of steps per iteration & 100 \\
    num of epochs & 5 \\
    gradient clipping & 1.0 \\
    adam epsilon & $1e-8$ \\

    \midrule[0.5pt]
    \multicolumn{2}{c}{\textbf{PPO}} \\
    clip range & 0.2 \\
    entropy coefficient & 0.01 \\
    discount factor $\gamma$ & 0.99 \\
    GAE balancing factor $\lambda$ & 0.95 \\
    desired KL-divergence & 0.01 \\
    actor and double critic NN & MLP, hidden units [512, 216, 128] \\

    \midrule[0.5pt]
    \multicolumn{2}{c}{\textbf{\beamdojo}} \\
    $w_1, w_2$ & 1.0, 0.25 \\
    
    \bottomrule[1.0pt]
    \end{tabular}
    \label{tab:hyperparameters}
\end{table}

\end{document}